\documentclass[lettersize,journal]{IEEEtran}
\usepackage{amsmath,amsfonts}
\usepackage{algorithmic}
\usepackage{array}
\usepackage[caption=false,font=normalsize,labelfont=sf,textfont=sf]{subfig}
\usepackage{textcomp}
\usepackage{stfloats}
\usepackage{url}
\usepackage{verbatim}
\usepackage{graphicx}
\usepackage{cite}

\def\BibTeX{{\rm B\kern-.05em{\sc i\kern-.025em b}\kern-.08em
    T\kern-.1667em\lower.7ex\hbox{E}\kern-.125emX}}
\usepackage{float}
\usepackage{pdfpages}
\usepackage{balance}
\usepackage{booktabs}
\usepackage{bm}
\usepackage{multirow}
\usepackage{makecell}
\usepackage{amssymb}
\usepackage{pgfplots}
\usepackage{multicol}

\usepackage{xcolor}
\usepackage{colortbl}
\usepackage{hhline}
\usepackage{hyperref}

\usepackage{tikz}
\usetikzlibrary{shapes.geometric, arrows.meta, positioning, shadows}
\usepackage{tikz-qtree}
\usetikzlibrary{decorations}
\usetikzlibrary{mindmap, shadows, trees}

\begin{document}
%\title{Deep Learning for Point Cloud Registration: A Comprehensive Survey and Taxonomy}
\title{Deep Learning-Based Point Cloud Registration: A Comprehensive Survey and Taxonomy}

%Yu-Xin Zhang, Jie Gui,~\IEEEmembership{Senior Member}, Baosheng Yu, Xiaofeng Cong, Xin Gong, Wenbing Tao, \\ Dacheng Tao,~\IEEEmembership{Fellow}, IEEE
\author{Yu-Xin Zhang, Jie Gui, Baosheng Yu, Xiaofeng Cong, Xin Gong, Wenbing Tao, Dacheng Tao
	\thanks{This work was supported in part by the grant of the National Science Foundation of China under Grant 62172090; Start-up Research Fund of Southeast University under Grant RF1028623097. We thank the Big Data Computing Center of Southeast University for providing the facility support on the numerical calculations. (Corresponding author: Jie Gui.)}
	\thanks{Yu-Xin Zhang, Jie Gui, Xiaofeng Cong, and Xin Gong are with the School of Cyber Science and Engineering, Southeast University, Nanjing 210000, China. Jie Gui is also with Purple Mountain Laboratories, Nanjing, 210000, China. (e-mail: yuxinzhang@seu.edu.cn, guijie@seu.edu.cn, cxf\_svip@163.com, xingong@seu.edu.cn)} 
    \thanks{Baosheng Yu is with the Lee Kong Chian School of Medicine, Nanyang Technological University, 308232, Singapore. (e-mail: baosheng.yu@ntu.edu.sg)}
    \thanks{Wenbing Tao is with the School of Artificial Intelligence and Automation, Huazhong University of Science and Technology, Wuhan, 430074, China. (e-mail: wenbingtao@hust.edu.cn)}
    \thanks{Dacheng Tao is with the College of Computing \& Data Science at Nanyang Technological University, \#32 Block N4 \#02a-014, 50 Nanyang Avenue, Singapore 639798 (email: dacheng.tao@gmail.com)}
  
  }
\markboth{Journal of \LaTeX\ Class Files,~Vol.~18, No.~9, September~2020}%
{How to Use the IEEEtran \LaTeX \ Templates}

\maketitle

\begin{abstract}
Point cloud registration involves determining a rigid transformation to align a source point cloud with a target point cloud. This alignment is fundamental in applications such as autonomous driving, robotics, and medical imaging, where precise spatial correspondence is essential. Deep learning has greatly advanced point cloud registration by providing robust and efficient methods that address the limitations of traditional approaches, including sensitivity to noise, outliers, and initialization. However, a well-constructed taxonomy for these methods is still lacking, making it difficult to systematically classify and compare the various approaches. In this paper, we present a comprehensive survey and taxonomy on deep learning-based point cloud registration (DL-PCR).
We begin with a formal description of the point cloud registration problem, followed by an overview of the datasets, evaluation metrics, and loss functions commonly used in DL-PCR. Next, we categorize existing DL-PCR methods into supervised and unsupervised approaches, as they focus on significantly different key aspects. For supervised DL-PCR methods, we organize the discussion based on key aspects, including the registration procedure, optimization strategy, learning paradigm, network enhancement, and integration with traditional methods; For unsupervised DL-PCR methods, we classify them into correspondence-based and correspondence-free approaches, depending on whether they require explicit identification of point-to-point correspondences. 
To facilitate a more comprehensive and fair comparison, we conduct quantitative evaluations of all recent state-of-the-art approaches, using a unified training setting and consistent data partitioning strategy. Lastly, we highlight the open challenges and discuss potential directions for future study. We hope this comprehensive survey will contribute to the advancement of DL-PCR by serving as a fundamental reference for researchers and practitioners in the field. A comprehensive collection of DL-PCR resources is available at \href{https://github.com/yxzhang15/PCR}{https://github.com/yxzhang15/PCR}.
\end{abstract}


\section{Introduction} \label{Sec-1}

Advancements in high-precision sensors, such as LiDAR and 3D cameras, have greatly improved the speed, accuracy, and cost-effectiveness of point cloud data collection. These sensors capture detailed 3D representations of the physical world, which are crucial for applications that require high spatial precision~\cite{guo2020deep, xiao2023unsupervised}. A key technique for processing this data is point cloud registration~\cite{besl1992method,jian2010robust}, which involves determining a rigid transformation to align a source point cloud with a target point cloud. This process enables the integration of point cloud data from diverse sources, including different sensors, viewpoints, or time frames, into a unified system. As a result, point cloud registration plays an essential role in various downstream applications, such as autonomous driving \cite{chen20203d}, 3D reconstruction \cite{huang2024surface}, pose estimation \cite{dang2022learning}, 3D localization \cite{elbaz20173d}, cultural heritage preservation \cite{gomes20143d}, robotics \cite{pomerleau2015review}, and medical imaging \cite{sinko20183d}.

Deep learning revolutionizes point cloud registration by automating feature extraction, increasing robustness to noise, enhancing computational efficiency, and offering adaptable solutions that effectively address a wide range of real-world challenges~\cite{lowens2024unsupervised, ao2023buffer, qin2023geotransformer, zhang2024constructing, chen2023full}. However, a well-constructed taxonomy for these methods is still lacking, making it difficult to systematically classify and compare the various approaches. To address this gap, we present a comprehensive survey and taxonomy of deep learning-based point cloud registration (DL-PCR). Our focus is on three key aspects: 1) establishing a clear and well-defined taxonomy, 2) providing a fair comparison of existing methods, and 3) identifying key challenges and proposing future research directions, as follows:

\begin{itemize}
    \item \textbf{Taxonomy.} We categorize DL-PCR methods into supervised and unsupervised approaches, as they focus on significantly different key aspects. Supervised methods rely on labeled training data to learn the mapping between source and target point clouds. These methods typically aim for higher accuracy by utilizing supervisory signals or ground-truth correspondences during the learning process. In contrast, unsupervised methods do not require labeled data. Instead, they leverage the intrinsic characteristics of the point clouds or the registration process itself, offering greater flexibility and scalability. This makes unsupervised methods particularly useful in real-world scenarios where labeled data may be scarce or unavailable. The fundamental differences in methodology lead to distinct approaches for addressing key challenges, such as robustness to noise, generalization to new environments, and computational efficiency.
    Therefore, for supervised DL-PCR methods, we organize the discussion based on key aspects, including the registration procedure, optimization strategy, learning paradigm, network enhancements, and integration with traditional methods; For unsupervised DL-PCR methods, we classify them into correspondence-based and correspondence-free approaches, depending on whether they require explicit identification of point-to-point correspondences. \\
    
    \item \textbf{Comparison.} The rapid advancement of deep learning has led to the development of numerous methods and a wide range of benchmark datasets for DL-PCR. However, these advancements also present challenges in making comprehensive and fair comparisons, as the methods often differ in their training strategies and data processing techniques. To provide a clearer understanding of the performance of various DL-PCR methods, we conduct quantitative evaluations of recent state-of-the-art approaches, using a unified training setting and consistent data partitioning strategy. These comparisons not only enhance the understanding of different DL-PCR methods but also help better identify key challenges that remain in this field.\\
    
    \item \textbf{Challenge \& Direction.} By establishing a clear taxonomy for comprehensive review and employing unified settings for quantitative evaluations, we can more effectively identify key challenges in DL-PCR and explore potential directions for future research. Specifically, several key areas that warrant further exploration include the development of more realistic and complex data generation methods, the incorporation of complementary multimodal information, the integration of large language models, and a greater focus on semi-supervised and unsupervised learning techniques. A thorough investigation of these topics will be essential for advancing DL-PCR.
\end{itemize}

\begin{table*}[t]
	\centering
	\caption{A summary of various datasets used for DL-PCR.}
    \renewcommand{\arraystretch}{1}
	\begin{tabular}{l|ccccccc}
		\toprule
		\textbf{Dataset} & \textbf{Year} & \textbf{Type} & \textbf{Classes} &\textbf{Capacity} &\textbf{Level}& \textbf{Scenarios}&\textbf{Representation}  \\	
		\midrule
		ETH \cite{pomerleau2012challenging} & 2012 & Real-World & 8&276&Scene-level&Indoor \& Outdoor&Point Clouds\\
		KITTI \cite{geiger2012we} &2012 & Real-World  &8 &22&Scene-level&Outdoor&Point Clouds\\
		ICL-NUIM \cite{choi2015robust} &2015 &Synthetic &2& 8 &Scene-level&Indoor &Depth Images\\
        ModelNet40 \cite{wu20153d} &2015 &Synthetic & 40&12311&Object-level&-&Mesh\\
		ShapeNetCore \cite{chang2015shapenet} &2015 & Synthetic & 55&51190&Object-level&-&Mesh\\
		RedWood \cite{choi2016large} &2016 &Real-World &9&1781  &Object-level&-& Depth Images\\  
		3DMatch \cite{zeng20173dmatch} &2017 & Synthetic \& Real-World &-& 62&Scene-level&Indoor \& Outdoor&Point Clouds\\	
		Oxford RobotCar \cite{maddern20171} &2017 & Real-World &-& 22703&Scene-level&Outdoor&Point Clouds\\
		ScanObjectNN \cite{uy2019revisiting}&2019 &Real-World  & 15&2902&Object-level&-&Point Clouds\\
		WHU-TLS \cite{dong2020registration}&2020 & Real-World &11 &115&Scene-level &Outdoor&Point Clouds\\
		NuScenes \cite{caesar2020nuscenes} &2020 &Real-World&23&1000&Scene-level&Outdoor &Point Clouds\\
            MVP-RG \cite{pan2021variational} &2021 & Synthetic &-&7600 & Object-level&- &Point Clouds\\ 
		FlyingShapes \cite{chen2023sira} &2023 & Synthetic &-&200 & Scene-level&Indoor &Point Clouds\\ 
		\bottomrule
	\end{tabular}
	\label{dataset}		
	\centering
\end{table*}

\subsection{The Difference Between This Survey and Others}
On the one hand, the surveys by \cite{gu2020review, bellekens2014survey, tam2012registration} primarily focus on traditional point cloud registration methods, without addressing the advancements brought by deep learning techniques; On the other hand, \cite{huang2021comprehensive, lyu2024rigid, zhao2024deep, huang2023cross, zhang2020deep} provide overviews of DL-PCR. However, these reviews, particularly \cite{huang2023cross, zhao2024deep}, predominantly concerntrate on specific subfields, such as low-overlap scenarios and cross-source registration. Moreover, none of these reviews offer an in-depth discussion on recent advancements in unsupervised methods. To address these gaps, \cite{ijcai2024p922} presents a comprehensive classification and analysis of recent DL-PCR methods, covering both supervised and unsupervised approaches. This work provides a more integrated and up-to-date overview of the field. This paper extends our previous work \cite{ijcai2024p922}, introducing several key improvements: (i) a more detailed taxonomy, (ii) a comprehensive overview that includes the most recent advancements, (iii) a comparison of various deep learning-based registration algorithms within a unified framework, and (iv) a reorganization of future trends to better reflect the ongoing advancements in deep learning techniques.

\subsection{A Guide for Reading This Survey}
Section \ref{Sec-2} provides the definition of DL-PCR, along with an overview of commonly used datasets, loss functions, and evaluation metrics. The taxonomy of both supervised and unsupervised methodologies is detailed in Sections \ref{Sec-3} and \ref{Sec-4}, respectively. Section \ref{Sec-5} presents experimental evaluations and analyzes existing approaches. Unresolved challenges and promising research directions are discussed in Section \ref{Sec-6}. Finally, Section \ref{Sec-7} summarizes the key points of this survey.

\section{Preliminaries} \label{Sec-2}

\subsection{Problem Definition}

Pairwise registration serves as a fundamental task in various DL-PCR applications, including multi-instance and multi-view registration. The objective is to determine the optimal rigid transformation $\bm{T}$, which consists of a rotation matrix $\bm{R} \in SO(3)$ and a translation vector $\bm{t} \in \mathbb{R}^3$, to align a source point cloud $\bm{X} \in \mathbb{R}^{N \times 3}$ with a target point cloud $\bm{Y} \in \mathbb{R}^{M \times 3}$ within a shared coordinate system. This process minimizes the alignment error between corresponding points, as formulated in the following equation:
\begin{equation}
    \bm{T}^{*} = \mathop{\text{argmin}}_{\bm{T}} \sum_{p=1}^P \|\bm{T}(\bm{x}_p) - \bm{y}_p\|^2,
\end{equation}
where $\bm{x}_p \in \bm{X}, \bm{y}_p \in \bm{Y}$ represent corresponding points from the source and target point clouds, respectively, and $P$ denotes the total number of such point pairs. 

In multi-instance registration, the target point cloud $\bm{Y}^{\prime}$ is divided into $K$ instances, denoted as $\{\bm{Y}^{\prime}_k\}_{k=1}^K$, where each instance $\bm{Y}^{\prime}_k$ represents a set of points that partially or fully sample the source point cloud $\bm{X}^{\prime}$. The objective is to compute a set of optimal rigid transformations, $\{(\bm{R}^{\prime}_k, \bm{t}^{\prime}_k)\}_{k=1}^K$, to align the source point cloud $\bm{X}^{\prime}$ with each instance $\bm{Y}^{\prime}_k$. In multi-view registration, the objective to align multiple point clouds $\{\bm{X}^{\prime\prime}_s\}_{s=1}^S$, each corresponding a distinct perspective or frame, into a unified coordinate system. This involves determining a set of global rigid transformations, $\{(\bm{R}^{\prime\prime}_s, \bm{t}^{\prime\prime}_s)\}_{s=1}^S$, that ensures consistent alignment across all views. Throughout this survey, unless explicitly stated otherwise, the discussion focuses on algorithms designed for pairwise registration.

\subsection{Datasets} 

This subsection reviews the datasets commonly used for training and evaluating DL-PCR methods. Table \ref{dataset} highlights their key characteristics, including release year, data type, number of classes, capacity, data level, scenarios, and data representation formats. Missing information is indicated by a dash (``-”). The datasets are further categorized based on the scenarios they represent, with a clear distinction between object-level and scene-level datasets.

\subsubsection{Object-level Dataset}
Point cloud data of individual objects, such as airplanes, computers, and cups, form an object-level dataset. The ModelNet40 dataset \cite{wu20153d} is a widely used object-level dataset that includes 40 categories, with each model originating from computer-aided design (CAD) and featuring multiple viewpoints and varying structures. The MVP registration (MVP-RG) dataset \cite{pan2021variational} provides detailed 3D data focused on various poses and occlusions, making it ideal for evaluating DL-PCR methods in challenging conditions. For real-world applications, the RedWood dataset \cite{choi2016large} offers high-resolution 3D scans of objects in cluttered environments, designed to evaluate algorithms under diverse object poses. The ShapeNetCore dataset \cite{chang2015shapenet} contains a comprehensive collection of object models, suitable for training and evaluating algorithms across a wide range of object types. Lastly, the ScanObjectNN dataset \cite{uy2019revisiting}, which includes 3D object scans from 15 categories, provides a realistic and challenging environment for evaluating registration techniques, particularly for the alignment of noisy and partial point clouds.

\subsubsection{Scene-level Dataset}
Scene-level datasets provide structural and environmental information about entire scenes, encompassing both indoor and outdoor scenes. The KITTI dataset \cite{geiger2012we} offers large-scale 3D point cloud data from urban, rural, and highway scenes, making it a valuable resource for evaluating  algorithms in dynamic driving environments. Its high-quality data, captured using LIDAR and cameras, has made it a popular choice for autonomous driving research. The ETH dataset \cite{pomerleau2012challenging} includes 3D scan data from urban and indoor environments, featuring complex point clouds that challenge registration algorithms under dynamic conditions. Similarly, the Oxford RobotCar dataset \cite{maddern20171} provides point cloud data collected in urban settings, while the WHU-TLS dataset \cite{dong2020registration} offers large-scale point clouds from outdoor environments. The ICL-NUIM dataset \cite{choi2015robust} provides realistic RGB-D images along with camera trajectories, facilitating the evaluation of algorithms in intricate indoor environments. The NuScenes dataset \cite{caesar2020nuscenes}, a multimodal collection, incorporates data from six cameras, five radars, and one LiDAR, making it well-suited for evaluating algorithms in complex and cluttered driving scenarios. The 3DMatch dataset \cite{zeng20173dmatch} combines both real-world depth images and synthetic data, incorporating several other datasets, such as SUN3D \cite{xiao2013sun3d}, 7Scenes \cite{shotton2013scene}, RGB-D Scenes v2 \cite{lai2014unsupervised}, BundleFusion \cite{dai2017bundlefusion}, and Analysis by Synthesis \cite{valentin2016learning}. As the first scene-level dataset composed entirely of synthetic data, FlyingShapes \cite{chen2023sira} aims to improve model performance on real-world datasets by engaging in the training process.
 
\subsection{Loss Functions} 

\subsubsection{Supervised DL-PCR} Several loss functions are commonly used in supervised DL-PCR,  including circle loss, overlap loss, point matching loss, and regression loss. Each of these loss functions is introduced below:
\begin{itemize}
    \item Circle loss optimizes feature descriptors by minimizing the distance between corresponding points while maximizing the distance between non-corresponding points \cite{sun2020circle, qin2023geotransformer, yu2023rotation, xie2024hecpg}. It enhances the ability of the descriptors to effectively distinguish between points that should match and those that should not.
    \item Overlap loss supervises the estimation of the overlap regions in the point cloud to be aligned by framing the problem as a binary classification task \cite{huang2021predator, xu2021omnet, zhang2024constructing}. It encourages the network to predict higher probabilities for points in overlapping regions and lower probabilities for non-overlapping points.  
    \item Point matching loss supervises the prediction of keypoint matching by classifying whether a point can be correctly matched with points in another point cloud \cite{qin2023geotransformer, huang2024consistency, yao2025pare}. It encourages the network to assign higher match scores to points that are correctly matched and lower scores to incorrect or ambiguous matches.
    \item Regression loss measures the error between the predicted transformation parameters (rotation matrix and translation vector) and the ground-truth transformation \cite{chen2023full, zhang2024constructing, wang2019prnet}. It minimizes the difference between the predicted and ground-truth transformation parameters, helping the network to accurately align the point clouds during the registration process.
\end{itemize}

\subsubsection{Unsupervised DL-PCR} In unsupervised DL-PCR, loss functions optimize registration results by measuring geometric differences and structural consistency between the point clouds to be registered. These loss functions refine the network's transformation predictions without relying on annotated supervision. The most commonly used discrepancy loss is Chamfer Distance, which calculates the distance from each point in the source point cloud to the nearest point in the target point cloud, summing these distances to evaluate the alignment \cite{lai2014unsupervised, huang2022unsupervised, mei2023unsupervised}. In addition to Chamfer Distance, other loss functions, such as Earth Mover's Distance \cite{sarode2019pcrnet}, Cosine Distance \cite{el2021bootstrap}, and L1 loss \cite{el2021unsupervisedr}, are also employed. These functions either assess the overall disparity between point clouds or measure similarity in feature space.

\definecolor{1}{HTML}{ABC64E}
\definecolor{2}{HTML}{c39398}
\definecolor{3}{HTML}{FBB463}
\definecolor{4}{HTML}{80B1D3}
\definecolor{5}{HTML}{F47F72}

\begin{figure*}[!t]
	\newcommand{\customfontsize}{\fontsize{8pt}{10pt}\selectfont}
	\tikzset{
		base/.style = {draw=black, thick, font=\customfontsize, rectangle},
		% root/.style = {base, minimum width=1.8 cm, minimum height=0.8cm, fill=none, decorate, decoration={zigzag}, align=center, text width=1cm},
        root/.style = {base, minimum width=1.7 cm, minimum height=0.8cm, fill=none, align=center, text width=1.17cm},
            process-1/.style = {base, minimum width=2.5 cm, minimum height=0.8cm, fill=none, rounded corners, align=center, text width=2.65 cm, draw=1},
            process-2/.style = {base, minimum width=2.5 cm, minimum height=0.8cm, fill=none, rounded corners, align=center, text width=2.65 cm, draw=2},
            process-3/.style = {base, minimum width=2.5 cm, minimum height=0.8cm, fill=none, rounded corners, align=center, text width=2.65 cm, draw=3},
            process-4/.style = {base, minimum width=2.5 cm, minimum height=0.8cm, fill=none, rounded corners, align=center, text width=2.65 cm, draw=4},
            process-5/.style = {base, minimum width=2.5 cm, minimum height=0.8cm, fill=none, rounded corners, align=center, text width=2.65 cm, draw=5},            
            process-11/.style = {base, minimum width=2.5 cm, minimum height=0.5cm, fill=none, rounded corners, align=center, text width=2.5 cm, draw=1},
            process-121/.style = {base, minimum width=2.5 cm, minimum height=0.7cm, fill=none, rounded corners, align=center, text width=2.5 cm, draw=1},
            process-122/.style = {base, minimum width=2.5 cm, minimum height=0.92cm, fill=none, rounded corners, align=center, text width=2.5 cm, draw=1},
            process-13/.style = {base, minimum width=2.5 cm, minimum height=1.68cm, fill=none, rounded corners, align=center, text width=2.5 cm, draw=1},            
            process-21/.style = {base, minimum width=2.5 cm, minimum height=0.56cm, fill=none, rounded corners, align=center, text width=2.5 cm, draw=2},
            process-221/.style = {base, minimum width=2.5 cm, minimum height=0.7cm, fill=none, rounded corners, align=center, text width=2.5 cm, draw=2},
            process-22/.style = {base, minimum width=2.5 cm, minimum height=0.92cm, fill=none, rounded corners, align=center, text width=2.5 cm, draw=2},
            process-31/.style = {base, minimum width=2.5 cm, minimum height=0.56cm, fill=none, rounded corners, align=center, text width=2.5 cm, draw=3},
            process-32/.style = {base, minimum width=2.5 cm, minimum height=0.6cm, fill=none, rounded corners, align=center, text width=2.5 cm, draw=3},
            process-41/.style = {base, minimum width=2.5 cm, minimum height=0.92cm, fill=none, rounded corners, align=center, text width=2.5 cm, draw=4},
            process-43/.style = {base, minimum width=2.5 cm, minimum height=1.32cm, fill=none, rounded corners, align=center, text width=2.5 cm, draw=4},
            process-51/.style = {base, minimum width=2.5 cm, minimum height=0.56cm, fill=none, rounded corners, align=center, text width=2.5 cm, draw=5},
            process-52/.style = {base, minimum width=2.5 cm, minimum height=0.8cm, fill=none, rounded corners, align=center, text width=2.5 cm, draw=5},
            process-1m1/.style = {base, minimum width=1.5 cm, minimum height=0.5cm, fill=1!10, rounded corners, align=center, text width=9.9 cm, dashed, draw=1},
            process-1m2/.style = {base, minimum width=1.5 cm, minimum height=0.8cm, fill=1!10, rounded corners, align=center, text width=9.9 cm, dashed, draw=1},
            process-1m3/.style = {base, minimum width=1.5 cm, minimum height=0.8cm, fill=1!10, rounded corners, align=center, text width=9.9 cm, dashed, draw=1},
            process-2m1/.style = {base, minimum width=1.5 cm, minimum height=0.5cm, fill=2!10, rounded corners, align=center, text width=9.9 cm, dashed, draw=2},
            process-2m2/.style = {base, minimum width=1.5 cm, minimum height=0.8cm, fill=2!10, rounded corners, align=center, text width=9.9 cm, dashed, draw=2},
            process-3m1/.style = {base, minimum width=1.5 cm, minimum height=0.4cm, fill=3!10, rounded corners, align=center, text width=9.9 cm, dashed, draw=3},
            process-3m2/.style = {base, minimum width=1.5 cm, minimum height=0.8cm, fill=3!10, rounded corners, align=center, text width=9.9 cm, dashed, draw=3},
            process-4m1/.style = {base, minimum width=1.5 cm, minimum height=0.5cm, fill=4!10, rounded corners, align=center, text width=9.9 cm, dashed, draw=4},
            process-4m3/.style = {base, minimum width=1.5 cm, minimum height=1cm, fill=4!10, rounded corners, align=center, text width=9.9 cm, dashed, draw=4},
            process-5m1/.style = {base, minimum width=1.5 cm, minimum height=0.5cm, fill=5!10, rounded corners, align=center, text width=9.9 cm, dashed, draw=5},
            process-5m2/.style = {base, minimum width=1.5 cm, minimum height=0.8cm, fill=5!10, rounded corners, align=center, text width=9.9 cm, dashed, draw=5},
		arrow/.style={black, line width=1pt}
	}
	
	\begin{tikzpicture}[node distance=0.2cm and 0.3cm, auto]
		
		% Nodes
		\node (Supervised) [root, minimum width=1, draw=black] {Supervised \\DL-PCR};
		\node (Registration Procedure) [process-1, right=of Supervised] {Registration Procedure};	
            %%%%%%%%%%%%%%%%%%%%%111
		\node (Descriptor Extraction) [process-13, right=0.3cm of Registration Procedure] {Descriptor Extraction};
    	\node (Descriptor Extraction Method) [process-1m3, right=0.2cm of Descriptor Extraction, minimum width=200] {\makecell{
        \{3DMatch \cite{zeng20173dmatch}\}$^{17}$ \{PPFNet \cite{deng2018ppfnet}\}$^{18}$ \{3DSmoothNet \cite{gojcic2019perfect}\}$^{19}$ \\ 
        \{D3Feat \cite{bai2020d3feat}\}$^{20}$
    \{SpinNet \cite{ao2021spinnet}, DIP \cite{poiesi2021distinctive}\}$^{21}$ 
    \{YOHO \cite{wang2022you}, GeDi \cite{poiesi2022learning}\}$^{22}$\\
    \{RoReg \cite{wang2023roreg}, SphereNet \cite{zhao2023spherenet}\}$^{23}$ 
    \{GMCNet \cite{pan2024robust}, DeepSGM \cite{liu2024deep}, 
    \\ RoCNet++ \cite{slimani2024rocnet++}, HA-TiNet \cite{zhao2024ha}\}$^{24}$
    \{PARE-Net \cite{yao2025pare}\}$^{25}$}};%
    
            \node (Overlap Prediction) [process-122, below=0.07cm of Descriptor Extraction] {Overlap Prediction};		
            \node (Overlap prediction Method) [process-1m2, right=0.2cm of Overlap Prediction,  minimum width=200]{\makecell{\{Predator \cite{huang2021predator}, OMNet \cite{xu2021omnet}\}$^{21}$, \{STORM \cite{wang2022storm}\}$^{22}$ \\
            \{RORNet \cite{wu2023rornet} Li et al. \cite{li2023unified}\}$^{23}$ \{MPC \cite{liu2024low}\}$^{24}$ 
            }};
        
            \node (Similarity Matrix  Optimization) [process-121, below=0.07cm of Overlap Prediction] {Similarity Matrix\\ Optimization};		
            \node (Similarity Matrix  Optimization Method) [process-1m2, right=0.2cm of Similarity Matrix Optimization,  minimum width=200]{\{PRNet \cite{wang2019prnet}\}$^{19}$ \{FIRE-Net \cite{wu2021feature}\}$^{21}$  \{OIF-PCR \cite{yang2022one}, SHM \cite{zhang2022end}\}$^{22}$};
    
            \node (Outlier Filtering) [process-122, below=0.07cm of Similarity Matrix  Optimization] {Outlier Filtering};		
            \node (Outlier Filtering Method) [process-1m2, right=0.2cm of Outlier Filtering,  minimum width=200]{\makecell{
            \{DHVR \cite{lee2021deep}, PointDSC \cite{bai2021pointdsc}\}$^{21}$ 
            \{DLF \cite{zhang2022partial}, SC$^2$-PCR \cite{chen2022sc2}\}$^{22}$ 
            \{SC$^2$-PCR++ \cite{chen2023sc}, \\ Hunter \cite{yao2023hunter}\}$^{23}$ 
            \{Yuan et al. \cite{yuan2024robust}, TEAR \cite{huang2024scalable}, MAC \cite{yang2024mac}, FastMAC \cite{zhang2024fastmac}\}$^{24}$ }};
    
            \node (Transformation Parameter Estimation) [process-122, below=0.07cm of Outlier Filtering] {Transformation \\Parameter Estimation};	
            \node (Transformation Parameter Estimation Method) [process-1m2, right=0.2cm of Transformation Parameter Estimation,  minimum width=200]{
            \{DetarNet \cite{chen2022detarnet}, FINet \cite{xu2022finet}, Zhang et al. \cite{zhang2022self}\}$^{22}$ \\
            \{MFANet \cite{yuan2024learning}, Q-REG \cite{jin2024q}\}$^{24}$ };
            
            \node (Overall) [process-122, below=0.07cm of Transformation Parameter Estimation] {Others};	
            \node (Overall Method) [process-1m2, right=0.2cm of Overall,  minimum width=200]{\makecell{
            \{MVDesc \cite{zhou2018learning}\}$^{18}$ \{DeepVCP \cite{lu2019deepvcp}\}$^{19}$ 
            \{Gojcic et al. \cite{gojcic2020learning}\}$^{20}$\\
            \{HRegNet \cite{lu2021hregnet}, RGM \cite{fu2021robust}\}$^{21}$
            \{BUFFER \cite{ao2023buffer}, Wang et al. \cite{wang2023robust}\}$^{23}$}};        
            %%%%%%%%%%%%%%%%%%%%%%%%%%%222		
		\node (Optimization Strategy) [process-2, below=5.42cm of Registration Procedure] {Optimization Strategy};	%
  
		\node (Gaussian Mixture Model) [process-21, right=0.3cm of Optimization Strategy] {GMM-Based};	%
    	\node (Gaussian Mixture Model Method) [process-2m1, right=0.2cm of Gaussian Mixture Model, minimum width=200] {
        \{DeepGMR \cite{yuan2020deepgmr}\}$^{22}$ \{OGMM \cite{mei2023overlap}, Chen et al. \cite{chen2023point}\}$^{23}$};%
     
		\node (Bayesian-based) [process-21, below=0.07cm of Gaussian Mixture Model] {Bayesian-Based};	%	
            \node (Bayesian-based Method) [process-2m1, right=0.2cm of Bayesian-based, minimum width=200] {\{VBReg \cite{jiang2023robust}\}$^{23}$};%
             
		\node (Diffusion Model) [process-22, below=0.07cm of Bayesian-based] {Diffusion-Based};	%
            \node (Diffusion Model Method) [process-2m2, right=0.2cm of Diffusion Model, minimum width=200] {
            \{Regiffusion \cite{xu2023point}, DiffusionPCR \cite{chen2023diffusionpcr}\}$^{23}$ \\
            \{PosDiffNet \cite{she2024posdiffnet}, PointDifformer \cite{she2024pointdifformer}, Jiang et al. \cite{jiang2024se}\}$^{24}$
            \{Diff-Reg \cite{wu2024diff}\}$^{25}$};%
            
            \node (Multimodal) [process-22, below=0.07cm of Diffusion Model] {Multimodality-Based};	%
            \node (Multimodal Method) [process-2m1, right=0.2cm of Multimodal, minimum width=200] {
            \{PCR-CG \cite{zhang2022pcr}, IMFNET \cite{huang2022imfnet}, ImLoveNet \cite{chen2022imlovenet}, GMF \cite{huang2022gmf}\}$^{22}$ \\
            \{IGReg \cite{xu2024igreg}\}$^{24}$ \{SemReg \cite{fung2025semreg}\}$^{25}$};%

        \node (Pre-trained) [process-21, below=0.07cm of Multimodal] {Pretrain-Based};	%
    	\node (Pre-trained Method) [process-2m1, right=0.2cm of Pre-trained, minimum width=200] {
        \{SIRA-PCR \cite{chen2023sira}, Yuan et al. \cite{yuan2023boosting}, ZeroReg \cite{wang2023zero}\}$^{23}$ 
        \{PointRegGPT \cite{chen2024pointreggpt}\}$^{25}$};%
            %%%%%%%%%%%%%%%%%%%%%%%%%%%333	
		\node (Learning Process) [process-3, below=3.2cm of Optimization Strategy] {Learning Paradigm};	%
  
		\node (Contrastive Learning) [process-31, right=0.3cm of Learning Process] {Contrastive Learning};	%
    	\node (Contrastive Learning Method) [process-3m1, right=0.2cm of Contrastive Learning, minimum width=200] {\makecell{
        \{PointCLM \cite{yuan2022pointclm}, SCRnet \cite{shao2022scrnet}\}$^{22}$ \{Liu et al. \cite{liu2023density}\}$^{23}$
        \{UMERegRobust \cite{haitman2025umeregrobust}\}$^{25}$}};%

		\node (Meta Learning) [process-31, below=0.07cm of Contrastive Learning] {Meta Learning};	%
    	\node (Meta Learning Method) [process-3m1, right=0.2cm of Meta Learning, minimum width=200] {\{Point-TTA \cite{hatem2023point}, 3D Meta-Registration \cite{wang20203d}\}$^{23}$};%

		\node (Reinforcement Learning) [process-32, below=0.07cm of Meta Learning] {Reinforcement Learning};	%
    	\node (Reinforcement Learning Method) [process-3m2, right=0.2cm of Reinforcement Learning, minimum width=200] {
        \{ReAgent \cite{bauer2021reagent}\}$^{21}$ \{Chen et al. \cite{chen2023point2}\}$^{23}$};%
            %%%%%%%%%%%%%%%%%%%%%%%%%%%444
		\node (Network Enhancement) [process-4, below=1.6cm of Learning Process] {Network Enhancement};	%
  
            \node (Attention Mechanism) [process-41, right=0.3cm of Network Enhancement] {Convolution-Based};	% 
    	\node (Attention Mechanism Method) [process-4m1, right=0.2cm of Attention Mechanism, minimum width=200] {
    \{FCGF \cite{choy2019fully}\}$^{19}$ \{DGR \cite{choy2020deep}, 3DRegNet \cite{pais20203dregnet}\}$^{20}$ \{PCAM \cite{cao2021pcam}\}$^{21}$ \{SACF-Net \cite{wu2023sacf}, Wu et al. \cite{wu2023accelerating}\}$^{23}$};

		\node (Transformer Module) [process-43, below=0.07cm of Attention Mechanism] {Transformer-Based};	%
    	\node (Transformer Module Method) [process-4m3, right=0.2cm of Transformer Module, minimum width=200] {
        \makecell{\{REGTR \cite{yew2022regtr}\}$^{22}$ \{GeoTransformer \cite{qin2023geotransformer}, RoITr \cite{yu2023rotation}, PEAL \cite{yu2023peal}, \\ RegFormer \cite{liu2023regformer}, DIT \cite{chen2023full}, EGST \cite{yuan2023egst}\}$^{23}$ \\
        \{SPEAL \cite{xiong2024speal}, NMCT \cite{wang2024neighborhood}, CAST \cite{huang2024consistency}, DCATr \cite{chen2024dynamic}, MIRETR \cite{yu2024learning}\}$^{24}$}, };%
            %%%%%%%%%%%%%%%%%%%%%%%%%%%555
            \node (Integration of Traditional Algorithms) [process-5, below=1.55cm of Network Enhancement] {Integration With Traditional Algorithms};	%
  
            \node (Iterative Closest Point) [process-52, right=0.3cm of Integration of Traditional Algorithms] {Iterative Closest Point};	%
    	\node (Iterative Closest Point Method) [process-5m2, right=0.2cm of Iterative Closest Point, minimum width=200] {
         \{DCP \cite{wang2019deep}\}$^{19}$ \{IDAM \cite{li2020iterative}\}$^{20}$ 
         \{DCPCR \cite{wiesmann2022dcpcr}, Global-PBNet \cite{zheng2022global}\}$^{22}$};%

		\node (Robust Point Matching) [process-52, below=0.07cm of Iterative Closest Point] {Robust Point Matching};	%
    	\node (Robust Point Matching Method) [process-5m2, right=0.2cm of Robust Point Matching, minimum width=200] {\{RPMNet \cite{yew2020rpm}\}$^{20}$};%

		\node (Lucas-Kanade) [process-51, below=0.07cm of Robust Point Matching] {Lucas-Kanade};	%
    	\node (Lucas-Kanade Method) [process-5m1, right=0.2cm of Lucas-Kanade, minimum width=200] {\{PointNetLK \cite{aoki2019pointnetlk}\}$^{19}$ \{PointNetLK Revisited \cite{li2021pointnetlk}\}$^{21}$};%
            %%%%%%%%%%%%%%%%%%%%%%%%%%%
  
		% Coordinate for the right end of the arrow lines
		\coordinate (RightCoordS) at ([xshift=0.88cm]Supervised); % Adjust the xshift value as needed
		\coordinate (BranchPoint) at ([xshift=-0cm]Supervised.east);
		
		\coordinate (RightCoordDE) at ([xshift=1.6cm]Registration Procedure); % Adjust the xshift value as needed
		\coordinate (BranchPointDE) at ([xshift=0cm]Registration Procedure.east);
		
		\coordinate (RightCoordCS) at ([xshift=1.6cm]Optimization Strategy); % Adjust the xshift value as needed
		\coordinate (BranchPointCS) at ([xshift=0cm]Optimization Strategy.east);

  		\coordinate (RightCoordLP) at ([xshift=1.6cm]Learning Process); % Adjust the xshift value as needed
		\coordinate (BranchPointLP) at ([xshift=0cm]Learning Process.east);

    	\coordinate (RightCoordNE) at ([xshift=1.6cm]Network Enhancement); % Adjust the xshift value as needed
		\coordinate (BranchPointNE) at ([xshift=0cm]Network Enhancement.east);

      	\coordinate (RightCoordTA) at ([xshift=1.6cm]Integration of Traditional Algorithms); % Adjust the xshift value as needed
		\coordinate (BranchPointTA) at ([xshift=0cm]Integration of Traditional Algorithms.east);		
		%%%%%%%%%%%%%%%%%%%%%%%%%%%%%%%%%%%%111
		\draw [arrow] (Supervised.east) -- (BranchPoint);		
		\draw [arrow] (BranchPoint) -- (RightCoordS) |- (Registration Procedure);
		\draw [arrow] (BranchPointDE) -- (RightCoordDE) |- (Descriptor Extraction);
            \draw [arrow] (BranchPointDE) -- (RightCoordDE) |- (Overlap Prediction);
		\draw [arrow] (BranchPointDE) -- (RightCoordDE) |- (Similarity Matrix Optimization);
		\draw [arrow] (BranchPointDE) -- (RightCoordDE) |- (Outlier Filtering);
		\draw [arrow] (BranchPointDE) -- (RightCoordDE) |- (Transformation Parameter Estimation);
  		\draw [arrow] (BranchPointDE) -- (RightCoordDE) |- (Overall);

            \draw [arrow] (Descriptor Extraction) -- (Descriptor Extraction Method);
		\draw [arrow] (Overlap Prediction) -- (Overlap prediction Method);		
            \draw [arrow] (Similarity Matrix Optimization) -- (Similarity Matrix Optimization Method);
		\draw [arrow] (Outlier Filtering) -- (Outlier Filtering Method);
            \draw [arrow] (Transformation Parameter Estimation) -- (Transformation Parameter Estimation Method);
		\draw [arrow] (Overall) -- (Overall Method);
		%%%%%%%%%%%%%%%%%%%%%%%%%%%%%%%%%%%%%%%%222
		\draw [arrow] (BranchPoint) -- (RightCoordS) |- (Optimization Strategy);
		\draw [arrow] (BranchPointCS) -- (RightCoordCS) |- (Gaussian Mixture Model);
		\draw [arrow] (BranchPointCS) -- (RightCoordCS) |- (Bayesian-based);
		\draw [arrow] (BranchPointCS) -- (RightCoordCS) |- (Diffusion Model);
		\draw [arrow] (BranchPointCS) -- (RightCoordCS) |- (Multimodal);
		\draw [arrow] (BranchPointCS) -- (RightCoordCS) |- (Pre-trained);
  
		\draw [arrow] (Gaussian Mixture Model) -- (Gaussian Mixture Model Method);	
            \draw [arrow] (Bayesian-based) -- (Bayesian-based Method);
            \draw [arrow] (Diffusion Model) -- (Diffusion Model Method);
            \draw [arrow] (Multimodal) -- (Multimodal Method);
            
		%%%%%%%%%%%%%%%%%%%%%%%%%%%%%%%%%%%%%%%%333
		\draw [arrow] (BranchPoint) -- (RightCoordS) |- (Learning Process);
		\draw [arrow] (BranchPointLP) -- (RightCoordLP) |- (Contrastive Learning);
		\draw [arrow] (BranchPointLP) -- (RightCoordLP) |- (Meta Learning);
		\draw [arrow] (BranchPointLP) -- (RightCoordLP) |- (Reinforcement Learning);
  
		\draw [arrow] (Contrastive Learning) -- (Contrastive Learning Method);	
            \draw [arrow] (Meta Learning) -- (Meta Learning Method);
            \draw [arrow] (Reinforcement Learning) -- (Reinforcement Learning Method);		
		%%%%%%%%%%%%%%%%%%%%%%%%%%%%%%%%%%%%%%%%444
            \draw [arrow] (BranchPoint) -- (RightCoordS) |- (Network Enhancement);
		\draw [arrow] (BranchPointNE) -- (RightCoordNE) |- (Attention Mechanism);
		\draw [arrow] (BranchPointNE) -- (RightCoordNE) |- (Transformer Module);

		\draw [arrow] (Attention Mechanism) -- (Attention Mechanism Method);	
            \draw [arrow] (Transformer Module) -- (Transformer Module Method);
            \draw [arrow] (Pre-trained) -- (Pre-trained Method);		
		%%%%%%%%%%%%%%%%%%%%%%%%%%%%%%%%%%%%%%%%555
  		\draw [arrow] (BranchPoint) -- (RightCoordS) |- (Integration of Traditional Algorithms);
		\draw [arrow] (BranchPointTA) -- (RightCoordTA) |- (Iterative Closest Point);
		\draw [arrow] (BranchPointTA) -- (RightCoordTA) |- (Robust Point Matching);
		\draw [arrow] (BranchPointTA) -- (RightCoordTA) |- (Lucas-Kanade);
  
		\draw [arrow] (Iterative Closest Point) -- (Iterative Closest Point Method);	
            \draw [arrow] (Robust Point Matching) -- (Robust Point Matching Method);
            \draw [arrow] (Lucas-Kanade) -- (Lucas-Kanade Method);		
		%%%%%%%%%%%%%%%%%%%%%%%%%%%%%%%%%%%%%%%%end
  
	\end{tikzpicture}
	\caption{A taxonomy of supervised DL-PCR algorithms. Methods published in the same year are grouped in curly brackets, with the superscripts $^{17}$, $^{18}$, $^{19}$, $^{20}$, $^{21}$, $^{22}$, $^{23}$, $^{24}$, and $^{25}$ indicate the publication years 2017, 2018, 2019, 2020, 2021, 2022, 2023, 2024, and 2025 respectively. Within each curly bracket, the listed methods are not in any particular chronological order.} 
	\label{fig:spcr}
\end{figure*}

\subsection{Evaluation Metrics}

Evaluation metrics are essential for assessing performance, as they facilitate the comparison of results and guide the selection of optimal network structures and parameters. These metrics are classified into two main categories, as follows.

\subsubsection{Error Characterization}  

Metrics in this category focus on quantifying the accuracy of registration and transformation parameters. Root Mean Squared Error (RMSE) and Mean Squared Error (MSE) evaluate overall alignment accuracy by calculating point-wise deviations. Mean Isotropic Error (MIE) and Mean Absolute Error (MAE) assess isotropic and absolute alignment discrepancies, respectively, while Chamfer Distance captures point-level proximity. Additionally, the coefficient of determination (R²) measures the goodness of fit between point clouds, and Relative Rotation Error (RRE) and Relative Translation Error (RTE) quantify the precision of the estimated rotation matrix and translation vector.
    
\subsubsection{Performance Robustness}  
These metrics evaluate the algorithm’s resilience to challenging scenarios, such as noise, outliers, and partial overlaps, as well as the quality of feature correspondences. Feature Matching Recall (FMR) measures the proportion of correctly matched feature point pairs relative to the total ground-truth feature correspondences, reflecting the algorithm’s ability to identify accurate matches. Inlier Ratio (IR) assesses the fraction of inliers among all matches produced by the algorithm, offering insights into the precision of the matching process. Moreover, Registration Recall (RR) evaluates the overall registration quality by calculating the fraction of point cloud pairs whose transformation errors, both rotation and translation, fall below specified thresholds.

\begin{figure*}
	\centering
	\includegraphics[scale=0.62]{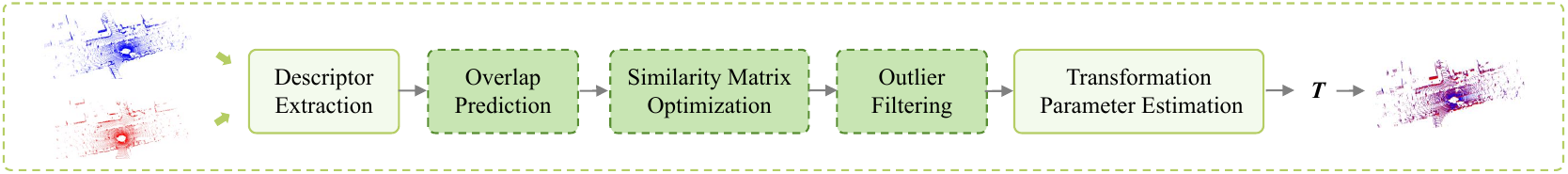}
	\caption{An overview of key registration procedures in supervised DL-PCR. Note that certain procedures, such as overlap prediction, similarity matrix optimization, and outlier filtering, may not be necessary for all registration methods.}
	\label{Fig:rp}
\end{figure*}

\section{Supervised DL-PCR} \label{Sec-3}

Supervised DL-PCR algorithms typically rely on various supervisory signals, such as ground-truth labels or transformation parameters, to guide the training process and optimize registration performance \cite{zhang2022vrnet, huang2024consistency, ao2023buffer}. To advance research in deep learning-based supervised methods, we categorize these algorithms into five main types based on the primary contributions of each study: (i) registration procedure (see Section \ref{Sec-rp}), (ii) optimization strategy (see Section \ref{Sec-os}), (iii) learning paradigm (see Section \ref{Sec-lp}), (iv) network enhancement (see Section \ref{Sec-ne}), and (v) integration of traditional algorithms (see Section \ref{Sec-cta}). A taxonomy of supervised DL-PCR algorithms is shown in Fig. \ref{fig:spcr}.

\subsection{Registration Procedure} \label{Sec-rp}
As shown in Fig. \ref{Fig:rp}, the supervised DL-PCR pipeline typically consists of five key procedures: descriptor extraction, overlap prediction, similarity matrix optimization, outlier filtering, and transformation parameter estimation. Each procedure plays a crucial role in enhancing registration accuracy. We categorize algorithms according to their focus within the registration process. Some algorithms optimize individual procedures, while others aim to jointly optimize multiple procedures to improve overall performance. The latter are grouped under the category of ``Others".

\subsubsection{Descriptor Extraction} Descriptor extraction serves to capture features with high discriminative power. To improve clarity, this category is further divided based on the type of input data, including point, patch, and voxel grid.

Firstly, \textit{point} is the basic unit of the point cloud, represented as a discrete, unconnected 3D entity. As a result, extracting descriptors from points often requires constructing complex local relationships. D3Feat \cite{bai2020d3feat} embeds the full convolutional network, which acquires the local information of the point cloud, into the joint learning framework to realize 3D local feature detection and description. GMCNet \cite{pan2024robust} utilizes rotation-invariant features and noise-resistant spatial coordinates to aggregate local features. The robust descriptors are subsequently generated by encoding the aggregated features through hierarchical graph networks and graph models. Additionally, geometric properties are captured by RoCNet++ \cite{slimani2024rocnet++} through a triangle-based local descriptor, which forms triangles between each point and its neighbors, leveraging the invariance of triangle angles under rigid transformations. While previous works focus on extracting local geometric descriptors, they often overlook the semantic context of scenes. To address this, DeepSGM \cite{liu2024deep} improves the robustness of large-scale DL-PCR by acquiring semantic instances through semantic segmentation and Euclidean clustering. Furthermore, a graph convolutional network is used to learn spatial and semantic features. To overcome rotation sensitivity, PARE-Net \cite{yao2025pare} proposes a position-aware rotation-equivariant network, which learns rotation-invariant features while maintaining distinctive geometric properties, ensuring efficient and stable registration.

Secondly, \textit{patch} reliably represents the local neighborhood structure in point clouds. As a pioneering work, 3DMatch \cite{zeng20173dmatch} laid the groundwork for extracting deep features from local patches. To incorporate rotation invariance, methods like PPFNet \cite{deng2018ppfnet} introduce global context awareness in descriptor learning, enabling single correspondences to consider other local features. YOHO \cite{wang2022you} further advances this by applying group-equivariant feature learning, which enables rotation-equivariant components that significantly reduce the search space for transformations. Based on YOHO, RoReg \cite{wang2023roreg} incorporates rotation-guided keypoint detection and rotation-coherent matching, which enhances correspondence quality. Other approaches, like DIP \cite{poiesi2021distinctive} and GeDi \cite{poiesi2022learning}, normalize the local reference frame and encode patch information into rotation-invariant descriptors through deep networks \cite{qi2017pointnet, qi2017pointnet++}. While both methods share a similar strategy, DIP focuses on rotation-invariant descriptors robust to clutter and occlusions, whereas GeDi further incorporates invariance to scale and point permutations, which allows better generalization across domains. Furthermore, HA-TiNet \cite{zhao2024ha} enhances rotation invariance by generating height-azimuth images and utilizing a ResNet-based backbone \cite{he2016deep} with a rotation-invariant layer. This method achieves full rotation invariance around all axes and provides a lightweight yet effective solution for local descriptor learning.

Thirdly, \textit{voxel grid} provides a uniform sampling of point clouds by using customizable grid size. 3DSmoothNet \cite{gojcic2019perfect} employs a voxelized smoothed density technique with fully convolutional layers to model local point cloud morphology, which leverages local density estimates for DL-PCR. In SpinNet \cite{ao2021spinnet}, the spatial point Transformer first maps the input point cloud into cylindrical space to achieve SO(2) equivariance. Subsequently, for a voxel at position $(j, k, l)$ on the $d$-th cylindrical feature map in the $s$-th layer, the 3D cylindrical convolution extracts features $\bm{F}$ by moving the convolution kernel with the following operation:
\begin{equation}
\bm{F}_{jkl}^{sd'} = \sum_{d=1}^{D} \sum_{r=1}^{R_s} \sum_{h=1}^{H_s} \sum_{a=1}^{A_s} w_{ryx}^{sd'd} \bm{F}_{(j+r)(k+h)(l+a)}^{(s-1)d},
\end{equation}
where $D$ denotes the number of input feature channels, $w_{ryx}^{sd'd}$ represents the learnable parameters, $R_s$ represents the size of the kernel along the radial dimension, $H_s$ and $A_s$ represents the kernel sizes along the height and width, respectively. Moreover, SphereNet \cite{zhao2023spherenet} learns noise-robust and generalizable descriptors by utilizing spherical interpolation, spherical integrity padding, and a spherical convolutional neural network. 

\subsubsection{Overlap Prediction} Methods focusing on enhancing this step estimate the overlap region between point clouds to be registered, and then search for correspondences within this region \cite{wu2022inenet, yu2023peal}. Predator \cite{huang2021predator} is the first model to introduce the concept of overlap prediction, which employs a graph neural network and an overlap attention module to enhance contextual relationships and predict overlap scores. To minimize the prediction error in overlapping regions, the overlap loss $L_{\text{overlap}}$ is defined as:
\begin{equation}
L_{\text{overlap}} = -\sum_{i=1}^{N} \left[o_i \log(\hat{o}_i) + (1 - o_i) \log(1 - \hat{o}_i) \right],
\end{equation}
where $N$ represents the total number of points, $o_i$ is the ground-truth label indicating whether the $i$-th point is within the overlap region (1 for within, 0 for outside), and $\hat{o}_i$ is the predicted probability for the $i$-th point. Based on this concept, OMNet \cite{xu2021omnet} introduces a mask prediction module for generating overlapping masks. Meanwhile, the intermediate layer of the module is combined with regression prediction, which can optimize mask generation and transform parameter estimation simultaneously. RORNet \cite{wu2023rornet} alleviates the problem of overlapping estimation errors by down-sampling and filtering out low similarity points. In addition, RORNet combines similarity-based and score-based methods in a two-branch structure to reduce noise sensitivity. STORM \cite{wang2022storm} integrates a differential sampling overlap prediction module within a dual Transformer \cite{vaswani2017attention} architecture to facilitate information exchange during the prediction stages. This module employs the Gumbel-Softmax \cite{jang2016categorical} to perform independent point sampling within the overlap region. In Li et al.\cite{li2023unified}, a bird's eye view (BEV) model is proposed that jointly learns 3D local features and overlap estimation. This BEV uses a sparse UNet-like network for feature characterization and a cross-attention module for overlapping region detection. Different from the algorithms above that directly predict the overlap region, MPC \cite{liu2024low} extends the overlap region through mutual reference, using low-overlap portions of point clouds to enhance the overlap. This approach does not rely on shaping prior knowledge, making it capable of handling various types of point cloud data, especially in low-overlap scenarios.

\subsubsection{Similarity Matrix Optimization} 
Each element of the similarity matrix represents the likelihood of a match between points in the point clouds to be registered. A probability function is commonly used to compute the similarity matrix, after which the maximum value in each row or column is identified to select the most likely point pairs~\cite{wang2019deep}. Although softmax is often employed as the probability function, it tends to generate a blurred correspondence map. To tackle this issue, various methods have been proposed to reduce the resulting ambiguity. PRNet \cite{wang2019prnet} proposes an iterative self-supervised framework for partial-to-partial registration, utilizing Gumbel-Softmax sampling \cite{jang2016categorical} to obtain an approximately differentiable matching matrix $\bm{m}$, defined as:
\begin{equation}
\bm{m}(\bm{x}_i, \bm{Y}) = \text{one\_hot}\left[\arg\max_j \text{softmax}\left(\frac{\bm{F}_Y \bm{F}_X^T + g_{ij}}{\lambda}\right)\right],
\end{equation}
where $\bm{F}_X$ and $\bm{F}_Y$ are the features of source and target point clouds extracted by DGCNN \cite{wang2019dynamic} and Transformer \cite{vaswani2017attention}, respectively. $g_{ij}$ is samples from a Gumbel(0, 1) distribution and $\lambda$ is a temperature parameter controlling the sharpness of the matching matrix. FIRE-Net \cite{wu2021feature} enhances feature interactions in multiple layers across the point cloud by extracting structural features and facilitating feature exchange, which in turn enables highly similar points to be recognized for easier matching. By considering a few correspondences, OIF-PCR \cite{yang2022one} proposes an effective position encoding strategy to normalize the point cloud point by point. The strategy is then incorporated into an iterative optimization process to progressively optimize the similarity matrix. SHM \cite{zhang2022end} adopts a soft-to-hard matching approach, where the enhanced Sinkhorn algorithm \cite{sinkhorn1964relationship} first computes a soft similarity matrix, which is then refined into the final permutation matrix. 

\subsubsection{Outlier Filtering} The goal of outlier filtering is to eliminate points that lack corresponding counterparts, known as outliers \cite{zhao2024sgor, li2024effective, han2024robust, ma2024pcgor}. As outliers can negatively affect the registration process, the removal of outliers is beneficial to maintain registration performance. DHVR \cite{lee2021deep} introduces a Hough Voting \cite{min2021convolutional} that places the predicted correspondences into a sparse transformation parameter space, improving the identification of inliers. Similarly, DLF \cite{zhang2022partial} employs a classifier with stacked order-aware modules to evaluate hypothesized outliers and verify the compatibility of inliers.

The above methods estimate outliers directly after feature extraction but usually ignore 3D spatial information due to over-reliance on multilayer perceptrons (MLP). In addition, they only evaluate feature pairs and do not take into account the consistency among inliers \cite{bai2021pointdsc}. To address these limitations, PointDSC \cite{bai2021pointdsc} explicitly incorporates spatial compatibility, which argues that not only should the relative distances between inliers in the point clouds remain consistent, but inliers within the same point cloud also share inherent spatial relationships. The  spatial compatibility can be represented by:
\begin{equation}
\bm{\beta}_{ij} = \left[1 - \frac{\bm{d}_{ij}^2}{\bm{\sigma}_d^2}\right]_{+}, \quad \bm{d}_{ij} = \left|\|\bm{x}_i - \bm{x}_j\| - \|\bm{y}_i - \bm{y}_j\|\right|,
\end{equation}
where $\bm{\beta}_{ij}$ measures the compatibility between point pairs $(\bm{x}_i, \bm{y}_i)$ and $(\bm{x}_j, \bm{y}_j)$, with $\bm{d}_{ij}$ as the length difference. The parameter $\bm{\sigma}_d$ adjusts sensitivity, and $[ \cdot ]_+$ ensures non-negativity. When $\bm{d}_{ij}$ exceeds $\bm{\sigma}_d$, $\bm{\beta}_{ij}$ becomes zero, indicating incompatibility. Moreover, SC$^2$-PCR \cite{chen2022sc2} introduces second-order spatial compatibility by converting the spatial compatibility matrix into a binary form and measuring similarity based on the count of mutually compatible points. This global compatibility approach improves early differentiation between inliers and outliers. Expanding on SC$^2$-PCR, SC$^2$-PCR++ \cite{chen2023sc} further refines the process by designing a feature and spatial consistency-constrained truncated Chamfer Distance metric, which enhances model hypothesis evaluation by increasing efficiency and reducing dependency on the assumed rate of true correspondences. Yuan et al. \cite{yuan2024robust} utilize spatial consistency to filter outliers but propose a joint integration strategy that leverages correspondences across different networks. Instead of recognizing inliers using first-order or second-order consistency as a prior in the above methods, Hunter \cite{yao2023hunter} learns higher-order consistency that is more robust to severe outliers. In TEAR \cite{huang2024scalable}, the outlier problem is tackled through a robust loss function based on truncated entry-wise absolute residuals, ensuring accurate registration even in highly contaminated datasets. Through constructing a compatibility graph and searching for maximal cliques, MAC \cite{zhang20233d, yang2024mac} filters out geometrically inconsistent outlier correspondences, generating accurate pose hypotheses to enhance registration precision. Building on MAC, FastMAC \cite{zhang2024fastmac} introduces graph signal processing and stochastic spectral sampling, prioritizing high-frequency nodes to further filter low-confidence outliers, thereby accelerating the hypothesis generation process.

\subsubsection{Transformation Parameter Estimation} The calculation of transformation parameters is the final step in DL-PCR. Random sample consensus (RANSAC) \cite{fischler1981random}, as a classic solving algorithm, is widely used in various DL-PCR methods. RANSAC is typically employed during the coarse registration phase to effectively reduce the impact of outliers, requiring a predetermined number of iterations to compute the transformation parameters. Additionally, some algorithms \cite{zhang2022point, chen2023full, zhang2024constructing} rely on singular value decomposition (SVD) \cite{arun1987least} to perform the calculation of transformation parameters. In contrast to RANSAC, SVD does not require iterative processing. It computes the optimal rotation and translation parameters by decomposing the covariance matrix $\bm{H}$, which $\bm{H}$ can be obtained by the following formula:  \begin{equation} 
\bm{H} = \sum_{i=1}^N (\bm{x}_i - \bar{\bm{x}})(\bm{y}_i - \bar{\bm{y}})^T, 
\end{equation} 
where $\bar{\bm{x}}$ and $\bar{\bm{y}}$ are the averages of the source and target point clouds, respectively. Apart from computing the transformation parameters using the traditional algorithms described above, several methods \cite{deng2018ppf, pais20203dregnet} have attempted to use convolutional neural networks to simultaneously solve for the rotation matrix and translation vector. These approaches have been validated by multiple models. However, solving for both transformation parameters together can lead to interference between the rotation and translation components \cite{chen2022detarnet}. To address this issue, DetarNet \cite{chen2022detarnet} employs a siamese network that decouples the rotation and translation parameters through a two-step process. First, a regression network is used to compute the translation vector, followed by the application of SVD to determine the rotation matrix. Additionally, FINet \cite{xu2022finet} enhances the information linkage between point clouds to be registered by leveraging both point-wise and global features. It utilizes a dual-branch structure with separate branches to predict the translation vector and rotation matrix. 

In addition to convolutional neural networks, some methods explore point cloud characteristics to solve for transformation parameters \cite{li2024dbdnet, zhang2022self}. MFANet \cite{yuan2024learning} introduces a framework using dual quaternion representation and multi-scale feature correlation to estimate transformations. By introducing a self-supervised framework, Zhang et al. \cite{zhang2022self} propose the concept of rigid transformation equivariance (RTE). RTE stipulates that when two point clouds (pose difference is $\bm{R}$, $\bm{t}$) undergo distinct rigid transformations $\bm{R}_1, \bm{t}_1$ and $\bm{R}_2, \bm{t}_2$, their new relative pose $\bm{\phi}^{\prime}$ relates to the original relative pose $\bm{\phi}$ can be obtained through the following relationship:
\begin{equation}
\bm{\phi}^{\prime} = \{ \bm{R}_2 \bm{R} \bm{R}_1^T, \bm{R}_2 \bm{t} + \bm{t}_2 - \bm{R}_2 \bm{R} \bm{R}_1^T \bm{t}_1 \}.
\end{equation}
Subsequently, this relationship is used to supervise the training of the network, which can be integrated into various registration frameworks. Q-REG \cite{jin2024q} utilizes quadric surface \cite{newman1993model} to estimate the rigid transform through a single correspondence. As well, Q-REG also employs an exhaustive search for robust estimation, enabling end-to-end training for correspondence matching and pose estimation. It does not depend on a specific matching method and is effective in improving performance during both inference and training.

\subsubsection{Others} \label{Sec-overall}
There are also some methods to achieve accurate registration by carefully designing multiple links in the registration procedure \cite{zhang2023pyrf, lu2021hregnet, ao2023buffer}. DeepVCP \cite{lu2019deepvcp}, an end-to-end registration network, which incorporates point weighting to estimate point saliency scores, facilitating the detection of keypoints. The $K$-nearest neighbors method is then utilized to create neighborhoods around these keypoints, followed by a permutation-invariant network that extracts more detailed descriptors. HRegNet \cite{lu2021hregnet} is a hierarchical network that utilizes geometric features, descriptors, and similarity measures obtained through bilateral and neighborhood consensus to establish correspondences between keypoints. The concepts of bilateral consensus and neighborhood consensus imply that, within the descriptor space, two correctly corresponding points should not only be nearest neighbors but also possess similar neighborhood structures. Through graph matching, RGM \cite{fu2021robust} not only considers the local geometry of each point when establishing correspondence relationships but also considers a larger range of structural and topological relationships. Furthermore, RGM trains the network using losses defined directly on the correspondences, promoting the network to learn better point-to-point correspondences. BUFFER \cite{ao2023buffer} designs a point-wise learner to enhance computational efficiency and feature representation capabilities by predicting keypoints and estimating point orientations. 

The above-mentioned papers focus on pairwise registration, there are some other studies focusing on multi-view registration tasks. MVDesc \cite{zhou2018learning} introduces a multi-view local descriptor designed to characterize 3D keypoints based on points captured from various perspectives. Following this, a robust matching technique is presented that rejects false correspondences through efficient belief propagation within a defined graphical model. Gojcic et al. \cite{gojcic2020learning} employ iteratively reweighted least squares (IRLS) as a global refinement method to preserve cycle consistency and reduce ambiguity in initial alignments. However, this approach relies on dense pairwise correspondences, which leads to a large computational overhead and an increase in outliers. This would make it difficult to make an accurate estimation of the correct pose for IRLS. To overcome these challenges, Wang et al. \cite{wang2023robust} propose a novel method focusing on learning reliable initialization techniques that take into account the overlap between multiple point cloud pairs. This strategy allows for the construction of sparse yet dependable pose graphs. Additionally, they incorporate a history reweighting function into the IRLS framework, enhancing its generalization and robustness.
 
\subsection{Optimization Strategy} \label{Sec-os}

\begin{figure}
	\centering
	\includegraphics[width=\linewidth]{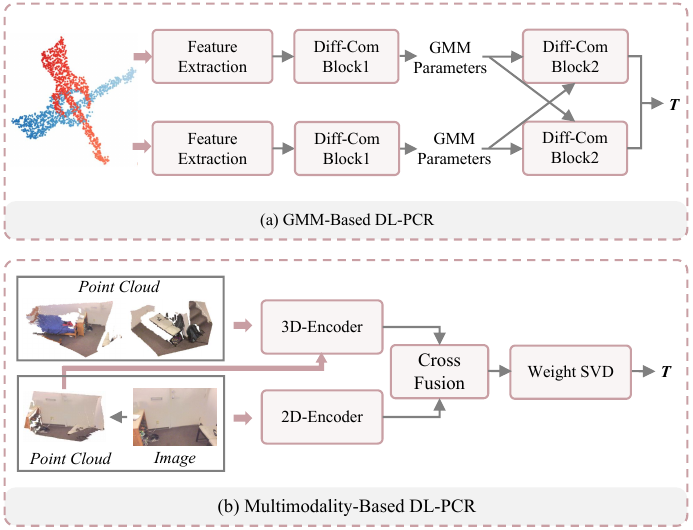}
	\caption{Illustration of DL-PCR methods using the following two optimization strategies: (a) GMM-based and (b) Multimodality-based.}
	\label{Fig:os}
\end{figure} 

These methods incorporate different optimization strategies into DL-PCR. Based on their underlying approaches, they are further classified into several categories: GMM-based, Bayesian-based, diffusion-based, multimodality-based, and pretrain-based techniques. The applications of each optimization strategy will be discussed.

\subsubsection{GMM-Based} As a commonly used probability model, the Gaussian Mixture Model (GMM) finds optimal alignments by integrating an expectation-maximization (EM) method into a maximum likelihood framework \cite{eckart2018hgmr, hertz2020pointgmm}. However, the EM process can be computationally intensive and potentially lead to incorrect data associations, especially in registrations with significant angular disparities \cite{yuan2020deepgmr}. To mitigate the problems highlighted above, a technique called deep Gaussian mixture registration (DeepGMR) \cite{yuan2020deepgmr} is proposed as shown in Fig. \ref{Fig:os}(a), which leverages the rigorously rotation-invariant processor \cite{chen2019clusternet} and correspondence network to search correspondences between points and GMM parameters. Specifically, two differentiable compute (Diff-Com) blocks are employed to estimate the optimal transformation parameters $\bm{T}^*$, with the optimization governed by
\begin{equation}
\bm{T}^* = \arg\min_{\bm{T}} \sum_{j=1}^J \frac{\hat{\bm{\pi}}_j}{\bm{\sigma}_j^2} \|\bm{T}(\hat{\bm{\mu}}_j) - \bm{\mu}_j\|^2,
\end{equation}
where ${J}$ represents the total number of components in the GMMs, $\hat{\bm{\pi}}_j$ and $\bm{\sigma}_j^2$ represent the source weights and target variances respectively, and $\hat{\bm{\mu}}_j$ and $\bm{\mu}_j$ represent the means of the source and target GMMs. In OGMM \cite{mei2023overlap}, predictions of the overlapping region between point clouds to be registered are utilized for GMM estimation, and the registration task is framed as minimizing the variance between the two GMMs. Chen et al. \cite{chen2023point} combined global and local information to formulate the GMM as a distribution that encompasses comprehensive representation capabilities.

\subsubsection{Bayesian-Based} Bayesian-based methods are statistical techniques for updating and improving estimates of event probabilities based on existing data. In VBReg \cite{jiang2023robust}, a non-local network architecture for DL-PCR is introduced, which utilizes variational Bayesian inference to learn non-local features. VBReg enables the modeling of Bayesian-driven long-range dependencies and facilitates the acquisition of discriminative feature representations for inlier/outlier.

\subsubsection{Diffusion-Based} Diffusion model is a machine learning technique based on generative stochastic processes \cite{sohl2015deep, ho2020denoising,luo2021diffusion, gong2024eadreg}, which generate data or optimize tasks by gradually introducing and eliminating noise. In DL-PCR, diffusion models are embedded at various stages, which provide key assistance in determining optimal alignment between source and target point clouds. PosDiffNet \cite{she2024posdiffnet} and PointDifformer \cite{she2024pointdifformer} utilize graph neural diffusion \cite{chamberlain2021grand} for assisting robust feature extraction. Specifically, PointDifformer uses graph partial differential equations and heat kernel signatures \cite{sun2009concise}, while PosDiffNet employs Beltrami flow and hierarchical matching to integrate positional information in large-scale registration. Regiffusion \cite{xu2023point} creatively employs the diffusion model to predict and generate overlapping regions between source and target point clouds before registration, offering a unique solution for scenarios with low or negative overlap rates. Diff-Reg \cite{wu2024diff} utilizes the denoising diffusion model to direct the search for the optimal matching matrix, focusing on efficiency and accuracy in finding better alignment solutions. Furthermore, Jiang et al. \cite{jiang2024se} apply the denoising diffusion process within the SE(3) space to refine the registration of point clouds for 6D object pose estimation, showcasing a novel approach by integrating linear Lie algebra SE(3) for precise spatial transformations. DiffusionPCR \cite{chen2023diffusionpcr} frames the registration task as a denoising process where noisy transformations are iteratively refined toward the ground-truth. In DiffusionPCR, the forward diffusion process is defined by
\begin{equation}
\bm{R}_t = {Slerp}(\bm{R}_{\text{quat}}^{\text{prior}}, \bm{R}_{\text{quat}}^{\text{gt}}; \sqrt{\alpha_t}), 
\end{equation}
\begin{equation}
\quad \bm{t}_t = (1-\sqrt{\alpha_t}) \cdot \bm{t}^{\text{prior}} + \sqrt{\alpha_t} \cdot \bm{t}^{\text{gt}}, 
\end{equation}
where $Slerp(\cdot,\cdot;\cdot)$ is the spherical linear interpolation~\cite{shoemake1985animating}, $\bm{R}_t$ and $\bm{t}_t$ are the rotation and translation at time $t$, and $\alpha_t$ controls the interpolation between the prior and ground-truth transformations, ensuring a smooth transition. The quaternion forms of the prior and ground-truth rotations are denoted by $\bm{R}_{\text{quat}}^{\text{prior}}$ and $\bm{R}_{\text{quat}}^{\text{gt}}$, respectively, while $\bm{t}^{\text{prior}}$ and $\bm{t}^{\text{gt}}$ refer to the prior and ground-truth translations.

% \subsubsection{Acceleration-Based} Sugiura et al. \cite{sugiura2023efficient} implement a dedicated accelerator IP core on a mid-range FPGA \cite{kosuge2020soc} and optimize it for feature extraction, improving performance without reducing generalization ability and accuracy. In addition, Sugiura et al. also research and develop an efficient accelerator for low-cost FPGA \cite{sugiura2024fpga}, achieving high registration speed and accuracy, and is suitable for resource-constrained computing platforms. Wu et al. \cite{wu2023accelerating} introduce a graph-based feature extraction module that simultaneously aggregates inter- and intra-contexts of the point clouds to be registered. Moreover, a sparse convolution-based feature refinement module is proposed that enhances the feature differences of dissimilar structures, enabling faster and more accurate registration in low-overlap scenarios.

\subsubsection{Multimodality-Based} Current multimodal DL-PCR algorithms aim to enhance the structural information of the original point cloud by integrating texture details extracted from images, which are beneficial for accurate representation and analysis \cite{xie2023cross, chen2022imlovenet, xu2024igreg}. PCR-CG \cite{zhang2022pcr} utilizes 2D image matching to create 2D correspondences, which are then mapped onto point clouds through a 2D-to-3D projection module to identify overlapping regions. Similarly, ImLoveNet \cite{chen2022imlovenet} leverages images to improve overlap region predictions, which uses cross-fusion technology to merge 3D features extracted from point clouds with 3D features simulated from 2D image-derived data, as shown in Fig. \ref{Fig:os}(b). IMFNET \cite{huang2022imfnet} introduces an interpretable module that explains the impact of original points on the final descriptor, thereby improving both transparency and effectiveness. By combining image information with point features and selectively fusing geometric consistency in reliably salient regions, IGReg \cite{xu2024igreg} improves the accuracy of alignment effectively in challenging scenarios including repetitive patterns and low geometric regions. Additionally, GMF \cite{huang2022gmf} combines the structural information from point clouds with texture information from images, allowing for effective outlier rejection. The aforementioned multimodal algorithms focus on enhancing registration accuracy through image data but lack specific supervision for optimizing 2D feature extraction, resulting in suboptimal matching performance \cite{fung2025semreg}. To address this issue, SemReg \cite{fung2025semreg} is proposed, which introduces a Gaussian mixture semantic prior to fusing 2D semantic features from images, revealing the semantic correlations between point cloud pairs. Furthermore, a semantics-aware focal loss is designed, which further optimizes the 2D feature extraction process that allows the model to better focus on semantic information.

\subsubsection{Pretrain-Based} Pre-training involves training the model on a large-scale dataset in advance, which allows the model to learn general features that provide a strong initialization for downstream tasks \cite{zheng2024point, yamada2022point}. In SIRA-PCR \cite{chen2023sira}, the model is pre-trained on the constructed large-scale synthetic dataset FlyingShapes to acquire general features. Meanwhile, a simulation-to-real adaptation process is employed to narrow the domain gap between synthetic and real-world point clouds. By transferring the knowledge of pre-trained multimodal models into a new point cloud descriptor neural network, Yuan et al. \cite{yuan2023boosting} can only require point cloud data during inference and achieve performance comparable to multimodal methods. ZeroReg \cite{wang2023zero} integrates pre-trained 2D visual features with 3D geometric information to achieve deep multimodal fusion, thereby enhancing the effectiveness of DL-PCR. PointRegGPT \cite{chen2024pointreggpt} utilizes the diffusion model in the pre-training phase to generate high-quality training data from the depth map. It enhances the realism and consistency of the data by re-projecting and correcting the depth map so that more precise and applicable training resources are provided.

\begin{figure}[!t]
	\centering
	\includegraphics[width=\linewidth]{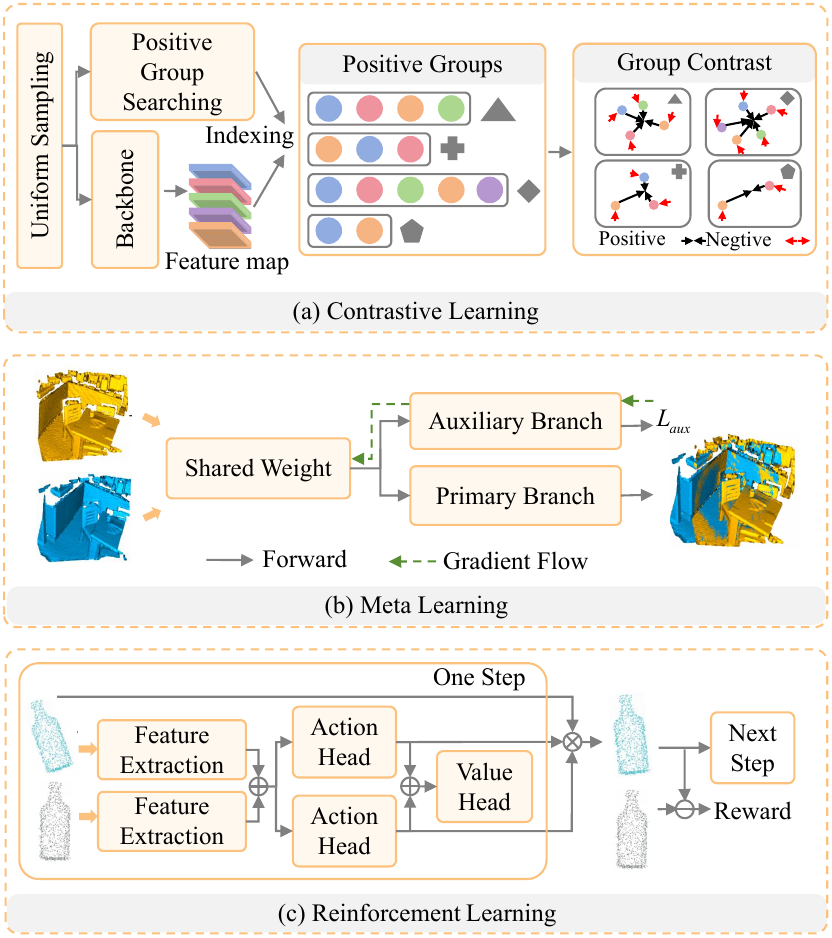}
	\caption{Illustration of DL-PCR methods using different learning paradigms: (a) contrastive learning, (b) meta learning, and (c) reinforcement learning.}
	\label{Fig:lp}
\end{figure} 

\subsection{Learning Paradigm} \label{Sec-lp}

Fig.~\ref{Fig:lp} illustrates the different learning paradigms used by DL-PCR methods: contrastive learning, meta-learning, and reinforcement learning. These learning paradigms aim to enhance feature representation, improve model generalization, and optimize the registration process through reward-based mechanisms, respectively.

\subsubsection{Contrastive Learning} Contrastive learning has emerged as a powerful tool in DL-PCR, enhancing feature representations and registration accuracy through various innovative approaches. In PointCLM \cite{yuan2022pointclm}, contrast learning is used to obtain a well-distributed depth representation $\bm{f}$ of the hypothesized counterparts. Then, these depth representations are used to prune outliers and cluster in-points, thus improving the multi-instance DL-PCR performance. The contrastive loss function $L_{cl}$ used in PointCLM is given by
\begin{equation}
\begin{split}
L_{cl} = \sum_{(i,j) \in O} \left[\text{E}(\bm{f}_i, \bm{f}_j) - \tau_o\right]^2_+ / |O| \\
  + \sum_{(i,j) \in N} \left[\tau_n - \text{E}(\bm{f}_i, \bm{f}_j)\right]^2_+ / |N|,
\end{split}
\end{equation}
Where $\text{E}$ represents the Euclidean Distance, $\tau_o$ and $\tau_n$ are margin thresholds for positive and negative pairs to prevent overfitting, $O$ and $N$ are the sets of positive and negative pairs, respectively, and $|O|$ and $|N|$ are the number of point pairs in $O$ and $N$. SCRnet \cite{shao2022scrnet} leverages contrastive learning and spatial consistency to guide the network for more accurate registration by emphasizing spatially consistent features. In addition, Liu et al. \cite{liu2023density} propose intergroup contrastive learning (as shown in Fig.~\ref{Fig:lp}(a)), which addresses the density mismatch problem in long-range outdoor DL-PCR by aligning multiple highly overlapping point clouds. It also ensures that positive samples are independently and identically distributed (i.i.d.), enabling density-invariant feature extraction. UMERegRobust \cite{haitman2025umeregrobust} integrates the transformation invariance property of universal manifold embedding \cite{efraim2022estimating, hagege2016universal} with the optimization strategy of contrastive learning. In this method, universal manifold embedding maps point cloud data into a low-dimensional subspace, providing a representation that is invariant to transformations. Meanwhile, contrastive learning optimizes the embedded features of the point cloud, ensuring that the model maintains consistency across transformed data.

\subsubsection{Meta Learning} The role of meta-learning in the DL-PCR task is to enhance generalization. As shown in Fig. \ref{Fig:lp}(b), Point-TTA \cite{hatem2023point} employs a test-time adaptation approach that updates the model parameters during inference using self-supervised auxiliary tasks, including point cloud reconstruction, feature learning, and correspondence classification, to adapt to unseen data distributions and improve registration performance. During the training phase, a meta-auxiliary learning framework is employed to optimize the model parameters $\theta$ using the following formula:
\begin{equation}
 \theta' = \theta - \gamma \nabla_\theta L_{\text{aux}}(\bm{X}, \bm{Y}, \theta),   
\end{equation}
where $L_{\text{aux}}$ denotes the loss function of the auxiliary tasks, and $\gamma$ is the learning rate. Another approach, 3D Meta-Registration \cite{wang20203d}, uses a 3D registration learner to focus on completing a specific registration task and a 3D registration meta-learner to provide optimal parameter update through task distribution-based training. With this synergistic approach, the parameters of the learner can be dynamically tuned to quickly adapt to new registration tasks.

\subsubsection{Reinforcement Learning} Reinforcement learning is applied to DL-PCR task to improve the registration performance. As shown in Fig. \ref{Fig:lp}(c), ReAgent \cite{bauer2021reagent} models DL-PCR as a reinforcement learning problem, where an agent interacts with the environment to optimize the registration process. The state of the agent is the current source point cloud $\bm{X}_i'$, and the action is the transformation applied to minimize the distance to the true source point cloud $\bm{X}$. The reward function in ReAgent is defined as:
\begin{equation}
r = 
\begin{cases} 
\varepsilon_+, & \text{if  D}(\bm{X}_i', \bm{X}) < \text{D}(\bm{X}_{i-1}', \bm{X}), \\
-\varepsilon_0, & \text{if  D}(\bm{X}_i', \bm{X}) = \text{D}(\bm{X}_{i-1}', \bm{X}), \\
-\varepsilon_-, & \text{if  D}(\bm{X}_i', \bm{X}) > \text{D}(\bm{X}_{i-1}', \bm{X}),
\end{cases}
\end{equation}
where $\bm{X}_{i-1}'$ is the source point cloud transformed by the transformation parameters obtained in the previous step and D is Chamfer Distance. The agent selects actions to maximize the reward, thereby gradually reducing registration errors and achieving precise alignment, which showcases the role of reinforcement learning in DL-PCR. Based on ReAgent, Chen et al. \cite{chen2023point2} enhance the state embedding with edge convolution for improved feature extraction and introduce a reward function that adjusts penalties over time, allowing for aggressive initial attempts followed by more conservative actions.

\subsection{Network Enhancement} \label{Sec-ne}
Techniques for enhancing networks include convolution-based operations \cite{choy20194d} and attention mechanisms \cite{niu2021review, vaswani2017attention}.

\begin{figure}
	\centering
	\includegraphics[width=\linewidth]{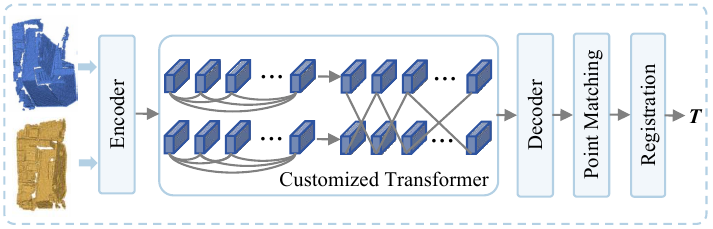}
	\caption{An example of Transformer-based DL-PCR.}
	\label{Fig:ne}
\end{figure} 

\subsubsection{Convolution-Based} These methods leverage advanced convolution strategies to model geometric structures in point clouds effectively. FCGF \cite{choy2019fully} introduces a fully convolutional geometric feature extraction approach based on sparse convolution, which improves feature learning efficiency and captures broader contextual information by extending the receptive field. DGR \cite{choy2020deep} employs a 6-dimensional convolutional network to predict the confidence of matched pairs. It integrates a weighted Procrustes method with a gradient optimization module to establish an end-to-end registration framework. 3DRegNet \cite{pais20203dregnet} combines ResNet \cite{he2016deep} with fully connected layers to classify matching point cloud pairs and regresses rigid transformation parameters for efficient and accurate registration. Wu et al.~\cite{wu2023accelerating} incorporates a sparse convolution module to expand the receptive field, enhancing feature discriminability and accelerating the estimation of transformation parameters. 
 Additionally, attention mechanisms have been adopted to dynamically adjust feature weights, focusing on key regions or features~\cite{zhao2023patch, wu2023sacf, wang2023ccag}. For example, PCAM \cite{cao2021pcam} uses cross-attention matrices (CAM) to enhance feature augmentation by simultaneously focusing on shallow geometric information and deep contextual details, generating more reliable matching features in overlapping regions. Inspired by PCAM, SACF-Net \cite{wu2023sacf} introduces a skip-attention-based correspondence filtering network that revisits encoder features at different resolutions to extract high-quality correspondences selectively. 

\subsubsection{Transformer-Based} Transformer architectures \cite{vaswani2017attention}, which leverage attention mechanisms to capture long-range dependencies and encode global relationships, have achieved remarkable success in DL-PCR \cite{yu2021cofinet, yu2024riga, li2022lepard, wang2024point, zhao2024lfa}. Fig. \ref{Fig:ne} illustrates a typical transformer-based DL-PCR framework. GeoTransformer \cite{qin2023geotransformer} encodes distance and angle information within the transformer framework to capture geometric structures in individual point clouds and ensure geometric consistency between point clouds during registration. Building on  GeoTransformer, several methods have introduced customized transformer networks for enhanced point cloud registration. For example, SPEAL \cite{xiong2024speal} improves upon GeoTransformer by incorporating skeleton-aware information to refine the registration process. RoITr \cite{yu2023rotation} introduces a rotation-invariant Transformer with an aggregation module, which helps extract discriminative, pose-agnostic descriptors while maintaining cross-frame position awareness. EGST \cite{yuan2023egst} develops an enhanced geometric structure Transformer that learns contextual features of geometric structures without relying on explicit position embeddings or additional feature exchange. PEAL \cite{yu2023peal} enhances registration accuracy by injecting overlap priors into the Transformer, specifically addressing low-overlap point cloud scenarios. NMCT \cite{wang2024neighborhood} improves local feature extraction with neighborhood position encoding and employs a multi-compound Transformer to facilitate local-global fusion and feature interaction. CAST \cite{huang2024consistency}presents an improved Transformer architecture that integrates consistency awareness, point-guided strategies, and a hierarchical design, addressing the challenges posed by complex registration scenarios. DCATr \cite{chen2024dynamic} proposes a dynamic cue-assisted Transformer that constrains attention to highly consistent regions and dynamically updates cue regions to improve feature distinctiveness and resolve matching ambiguities. 

Furthermore, several methods also directly use transformer networks to predict correspondences. REGTR \cite{yew2022regtr} employs a transformer-based architecture that integrates both self-attention and cross-attention mechanisms. This design enhances the network’s ability to extract meaningful features, enabling precise predictions of the probability that each point belongs to the overlap region. To mitigate false matches, RegFormer \cite{liu2023regformer} introduces a bijective association transformer, which regresses the initial transformation. This is followed by iterative refinement of shallow layers to recover the accurate transformation. To address matching and occlusion challenges in multi-instance DL-PCR, MIRETR \cite{yu2024learning} proposes an instance-aware geometric transformer. This model extracts instance-specific superpoint features and predicts coarse-level masks for each instance. An iterative optimization strategy is then applied to improve the reliability of the superpoint features and the accuracy of the masks, reducing feature contamination between instances.

\definecolor{11}{HTML}{FFD700}
\definecolor{22}{HTML}{77DCDD}

\begin{figure*}[!t]
	\newcommand{\customfontsize}{\fontsize{8pt}{10pt}\selectfont}
	\tikzset{
		base/.style = {draw=red, thick, font=\customfontsize, rectangle},
		root/.style = {base, minimum width=1.5 cm, minimum height=0.8 cm, fill=none, align=center, text width=1.5cm},
            process-purple/.style = {base, minimum width=2.65 cm, minimum height=0.8cm, fill=none, rounded corners, align=center, text width=2.65 cm, draw=11},
            process-red/.style = {base, minimum width=2.65 cm, minimum height=0.8cm, fill=none, rounded corners, align=center, text width=2.65 cm, draw=22},
		process-purple-2/.style = {base, minimum width=2.4 cm, minimum height=0.8cm, fill=none, rounded corners, align=center, text width=2.4 cm, draw=11},	
            process-purple-3/.style = {base, minimum width=2.4 cm, minimum height=0.84cm, fill=none, rounded corners, align=center, text width=2.4 cm, draw=11},
            process-red-2/.style = {base, minimum width=2.4 cm, minimum height=0.56cm, fill=none, rounded corners, align=center, text width=2.4 cm, draw=22},
            process-red-22/.style = {base, minimum width=2.4 cm, minimum height=0.9cm, fill=none, rounded corners, align=center, text width=2.4 cm, draw=22},
            process-red-3/.style = {base, minimum width=2.4 cm, minimum height=0.8cm, fill=none, rounded corners, align=center, text width=2.4 cm, draw=22},
            process-red-4/.style = {base, minimum width=2.4 cm, minimum height=0.84cm, fill=none, rounded corners, align=center, text width=2.4 cm, draw=22},
		process-purple-m1/.style = {base, minimum width=1.5 cm, minimum height=0.56cm, fill=11!10, rounded corners, draw=22, dashed, align=center, text width=9.8 cm, draw=11},	
            process-purple-m2/.style = {base, minimum width=1.5 cm, minimum height=0.8cm, fill=11!10, rounded corners, draw=22, dashed, align=center, text width=9.8 cm, draw=11},
            process-red-m1/.style = {base, minimum width=1.5 cm, minimum height=0.5cm, fill=22!10, rounded corners, align=center, text width=9.8 cm, dashed, draw=22},
            process-red-m2/.style = {base, minimum width=1.5 cm, minimum height=0.8cm, fill=22!10, rounded corners, align=center, text width=9.8 cm, dashed, draw=22},
		arrow/.style={black, line width=1pt}
	}
	
	\begin{tikzpicture}[node distance=0.25cm and 0.3cm, auto]	
		% Nodes
		\node (Unsupervised) [root, minimum width=1.5cm, draw=black] {Unsupervised\\DL-PCR}; 
            %%%%%%%%%%%%%%%%%%%%%111
		\node (Correspondence-free) [process-purple, right=0.25cm of Unsupervised] {Correspondence-Free};	
            
		\node (One-time Registration) [process-purple-2, right=0.25cm of Correspondence-free] {One-Stage Registration};
    	\node (One-time Registration Method) [process-purple-m2, right=0.2cm of One-time Registration, minimum width=200] {\{PPF-FoldNet \cite{deng2018ppf}\}$^{18}$ \{MS-SVConv \cite{horache20213d}, UPCR \cite{zhang2021representation}\}$^{21}$ \{UGMM \cite{huang2022unsupervised}\}$^{22}$};%
    
            \node (Iterative Registration) [process-purple-3, below=0.07cm of One-time Registration] {Iterative Registration};		
            \node (Iterative Registration Method) [process-purple-m1, right=0.2cm of Iterative Registration,  minimum width=200]{
            \{PCRNet \cite{sarode2019pcrnet}\}$^{19}$ \{Deep-3DAligner \cite{wang2020unsupervised}\}$^{20}$ \{Sun et al. \cite{sun2023research}\}$^{23}$ \{MMMI \cite{yuan2024MMIPCR}\}$^{24}$};      
            %%%%%%%%%%%%%%%%%%%%%%%%%%%222		
		\node (Correspondence-based) [process-red, below=1.09cm of Correspondence-free] {Correspondence-Based};	%
  
		\node (Enhanced by RGB-D) [process-red-22, right=0.25cm of Correspondence-based] {RGB-D};	%
            \node (Enhanced by RGB-D Method) [process-red-m1, right=0.2cm of Enhanced by RGB-D,  minimum width=200]{
            \{UnsupervisedR\&R \cite{el2021unsupervisedr}\}$^{21}$ \{LLT \cite{wang2022improving}\}$^{22}$ \\
\{PointMBF \cite{yuan2023pointmbf}\}$^{23}$  \{Yan et al. \cite{yan2024discriminative}, NeRF-UR \cite{yu2024nerf}\}$^{24}$};
     
		\node (Probability Model) [process-red-4, below=0.07cm of Enhanced by RGB-D] {Probability Model};	%	
            \node (Probability Model Method) [process-red-m1, right=0.2cm of Probability Model,  minimum width=200]{
          \{LatentCEM \cite{jiang2021planning}, CEMNet \cite{jiang2021sampling}\}$^{21}$ \\
          \{UGMM \cite{huang2022unsupervised}\}$^{22}$ 
          \{OBMNet \cite{shi2023overlap}, UDPReg \cite{mei2023unsupervised}\}$^{23}$};
             
		\node (Feature-based) [process-red-2, below=0.07cm of Probability Model] {Descriptor-Based};	%
            \node (Feature-based Method) [process-red-m1, right=0.2cm of Feature-based,  minimum width=200]{
            \{DeepUME \cite{lang2021deepume}, CorrNet3D \cite{zeng2021corrnet3d}\}$^{21}$ \{R-PointHop \cite{kadam2022r}\}$^{22}$ 
            \{GTINet \cite{jiang2024gtinet}\}$^{24}$};
  
		\node (Geometry Structure-based) [process-red-3, below=0.07cm of Feature-based] {Geometric Consistency-Based};	%	
            \node (Geometry Structure-based Method) [process-red-m2, right=0.2cm of Geometry Structure-based,  minimum width=200]{
            \{RIENet~\cite{shen2022reliable}\}$^{22}$ \{ICC \cite{yuan2024inlier}, RegiFormer \cite{zheng2024regiformer}, EYOC \cite{liu2024extend}, INTEGER \cite{xiong2024mining}\}$^{24}$};
            %%%%%%%%%%%%%%%%%%%%%%%%%%%
  
		% Coordinate for the right end of the arrow lines
		\coordinate (RightCoordS) at ([xshift=1.01cm]Unsupervised); % Adjust the xshift value as needed
		\coordinate (BranchPoint) at ([xshift=-0cm]Unsupervised.east);
		
		\coordinate (RightCoordDE) at ([xshift=1.6cm]Correspondence-free); % Adjust the xshift value as needed
		\coordinate (BranchPointDE) at ([xshift=0cm]Correspondence-free.east);
		
		\coordinate (RightCoordCS) at ([xshift=1.6cm]Correspondence-based); % Adjust the xshift value as needed
		\coordinate (BranchPointCS) at ([xshift=0cm]Correspondence-based.east);	
		%%%%%%%%%%%%%%%%%%%%%%%%%%%%%%%%%%%%111
		\draw [arrow] (Unsupervised.east) -- (BranchPoint);		
		\draw [arrow] (BranchPoint) -- (RightCoordS) |- (Correspondence-free);
		\draw [arrow] (BranchPointDE) -- (RightCoordDE) |- (One-time Registration);
            \draw [arrow] (BranchPointDE) -- (RightCoordDE) |- (Iterative Registration);

            \draw [arrow] (Iterative Registration) -- (Iterative Registration Method);
		\draw [arrow] (One-time Registration) -- (One-time Registration Method);		
		%%%%%%%%%%%%%%%%%%%%%%%%%%%%%%%%%%%%%%%%222
		\draw [arrow] (BranchPoint) -- (RightCoordS) |- (Correspondence-based);
		\draw [arrow] (BranchPointCS) -- (RightCoordCS) |- (Enhanced by RGB-D);
		\draw [arrow] (BranchPointCS) -- (RightCoordCS) |- (Probability Model);
		\draw [arrow] (BranchPointCS) -- (RightCoordCS) |- (Feature-based);
		\draw [arrow] (BranchPointCS) -- (RightCoordCS) |- (Geometry Structure-based);
  
		\draw [arrow] (Enhanced by RGB-D) -- (Enhanced by RGB-D Method);	
            \draw [arrow] (Probability Model) -- (Probability Model Method);
            \draw [arrow] (Feature-based) -- (Feature-based Method);
            \draw [arrow] (Geometry Structure-based) -- (Geometry Structure-based Method);
		%%%%%%%%%%%%%%%%%%%%%%%%%%%%%%%%%%%%%%%%end
  
	\end{tikzpicture}
	\caption{A taxonomy of unsupervised DL-PCR algorithms.}
	\label{fig:unpcr}
\end{figure*}

\subsection{Integration of Traditional Algorithms} \label{Sec-cta}

Traditional methods play a crucial role in supporting deep learning-based registration. Key examples of these methods include the integration of Iterative Closest Point (ICP), Robust Point Matching (RPM), and Lucas-Kanade (LK).

\subsubsection{Iterative Closest Point} As a classical method, ICP iteratively finds the nearest neighbor correspondences and estimates the rigid transformation~\cite{besl1992method, segal2009generalized}. Several DL-PCR algorithms combine the iterative optimization idea of ICP to optimize the registration process. DCPCR \cite{wiesmann2022dcpcr} introduces compressed encoders and attention mechanisms to improve the robustness of feature matching. Furthermore, DCP \cite{wang2019deep} uses attention mechanisms to predict soft correspondences between point clouds and employs a differentiable SVD layer to extract the rigid transformation. Contrasting with the above-discussed algorithms that iterate over the entire network, IDAM \cite{li2020iterative} distinctively positions the feature extraction component outside the iterative loop, which reduces the computational burden to a certain extent. Furthermore, it integrates distance information into the iterative network and incorporates a two-stage point elimination module. This design effectively filters out points and point pairs that are detrimental to the registration process. Global-PBNet \cite{zheng2022global} integrates deep learning for high-precision rough registration with ICP and branch-and-bound  optimization \cite{narendra1977branch} to achieve globally optimal alignment.

\subsubsection{Robust Point Matching} 
As a prominent work to address the limitations of ICP, RPM \cite{gold1998new} solves issues related to local minima and noise sensitivity, and it is also extended to a deep learning-based registration framework. RPMNet \cite{yew2020rpm} improves the classical RPM algorithm by learning hybrid features to compute the soft correspondence matrix $\bm{c}$ between the source and target point clouds:
\begin{equation}
\bm{c}_{ij} = e^{-\beta (\|\bm{F}_{\bm{x}_i} - \bm{F}_{\bm{y}_j}\|_2^2 - \alpha)},
\end{equation}
where $\bm{F}_{\bm{x}_i}$ and $\bm{F}_{\bm{y}_j}$ represent the learned features for points $\bm{x}_i$ and $\bm{y}_j$, respectively. The parameter $\beta$ controls the hardness of the matching, gradually converging during the iterative process, and $\alpha$ is used to reject outliers. The utilized differentiable Sinkhorn layer \cite{sinkhorn1964relationship} and annealing strategy reduce the dependence on initialization while enhancing robustness to noise and outliers, which enables the model to better handle partially visible point cloud data.

\subsubsection{Lucas-Kanade} The LK algorithm~\cite{lucas1981iterative} is a classical image registration method that iteratively optimizes rigid transformation parameters by minimizing the feature differences between a source and a target image. PointNetLK \cite{aoki2019pointnetlk} extends this concept to point cloud data by employing PointNet \cite{qi2017pointnet} to learn a feature mapping function $\phi$ that projects point clouds into a fixed-dimensional feature space. The registration process in PointNetLK aims to minimize the following objective function:
\begin{equation}
\arg \min_{\boldsymbol{\xi}} \|\phi(G(\boldsymbol{\xi}) \cdot \bm{X}) - \phi(\bm{Y})\|_2^2,
\end{equation}
where $G(\boldsymbol{\xi}) = \exp\left( \sum_{p=1}^{6} \boldsymbol{\xi}_{p} \boldsymbol{A}_{p} \right) \in \text{SE}(3)$, $\boldsymbol{\xi} \in \mathbb{R}^6$ represent exponential map twist parameters and $\boldsymbol{A}$ denote the generator matrices. This method iteratively updates the $\Delta \boldsymbol{\xi}$ to achieve registration. To enhance the generalization ability, PointNetLK Revisited \cite{li2021pointnetlk} improves upon the PointNetLK by introducing an analytic Jacobian matrix, addressing numerical instability issues and improving robustness. 

\begin{figure*}
	\centering
	\includegraphics[scale=0.75]{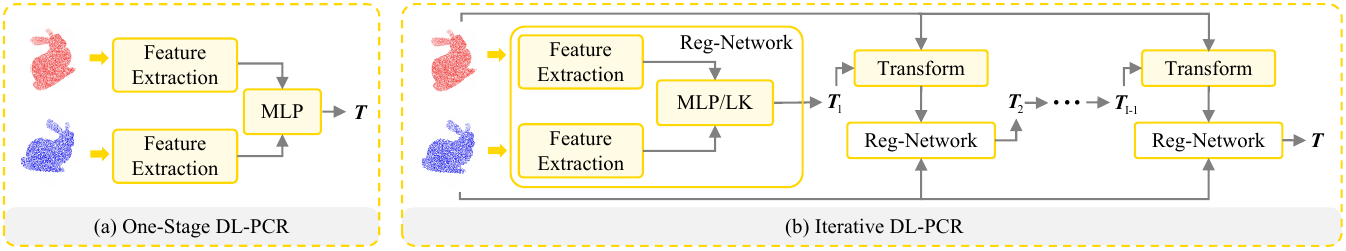}
	\caption{Pipeline visualization of (a) one-stage DL-PCR and (b) iterative DL-PCR in correspondence-free unsupervised methods.}
	\label{Fig:cfu}
\end{figure*}

\section{Unsupervised DL-PCR} \label{Sec-4}

Supervised DL-PCR algorithms have demonstrated promising results, but their effectiveness depends heavily on the availability of a large set of ground-truth transformations or correspondences to guide the training process. However, obtaining such annotated data in real-world scenarios is both challenging and costly, which limits the practical use of supervised registration algorithms. In contrast, unsupervised DL-PCR algorithms aim to address these limitations by eliminating the need for ground-truth labels. This section classifies unsupervised DL-PCR algorithms into two categories: correspondence-free and correspondence-based.

\subsection{Correspondence-Free}

Correspondence-free unsupervised registration methods begin by extracting global features from both the source and target point clouds. These features are then processed using a multi-layer perceptron (MLP) to regress the transformation parameters. We further categorize correspondence-free DL-PCR methods into two approaches: one-stage registration and iterative registration methods.
 
\subsubsection{One-Stage Registration} In the field of one-stage unsupervised registration methods, an early significant contribution is PPF-FoldNet \cite{deng2018ppf}, which starts by constructing four-dimensional (4D) point-pair features. These features are subsequently fed into an end-to-end architecture similar to a folded network as well as reconstructed using an encoder-decoder structure. The loss function in PPF-FoldNet involves comparing the Chamfer Distance (CD) between the 4D point pair features before and after reconstruction. The CD is a bidirectional metric that measures the discrepancy between two sets of data, which is commonly used in one-stage registration tasks. Specifically, the formula for CD is:
\begin{equation}
    \mathcal{L}\left(\bm{U}, \bm{V} \right)=  \frac{1}{|\bm{U}|} \sum\limits_{\bm{u} \in \bm{U}} \min\limits_{\bm{v} \in \bm{V}} \| \bm{u}-\bm{v} \|_2^2 
     + \frac{1}{|\bm{V}|} \sum\limits_{\bm{v} \in \bm{V}} \min\limits_{\bm{u} \in \bm{U}} \| \bm{v}-\bm{u} \|_2^2,
\end{equation}
where $\bm{U}$ and $\bm{V}$ could be two sets of points or their corresponding features, $|\bm{U}|$ and $|\bm{V}|$ are the number of points in $\bm{U}$ and $\bm{V}$. MS-SVConv \cite{horache20213d} combines multi-scale sparse voxel convolution and unsupervised transfer learning to enhance unsupervised registration. To learn relative poses that are essential for deriving transformation parameters, UPCR \cite{zhang2021representation} introduces dual point cloud representations: pose-invariant and pose-related. Moreover, the CD is also integrated to evaluate the discrepancy between the source point cloud and the target point cloud. UGMM \cite{huang2022unsupervised} estimates posterior probabilities through unsupervised learning, which uses the CD between Gaussian mixture models derived from the point clouds to be registered as the loss function.

\subsubsection{Iterative Registration} As shown in Fig. \ref{Fig:cfu}(b), the iterative registration method performs multiple iterations of the entire registration network (Reg-Network) to estimate the optimal solution. The source point cloud of each iteration is updated with the transformation parameters obtained in the previous iteration, while the target point cloud remains unchanged. Similar to the one-stage methods, Deep-3DAligner \cite{wang2020unsupervised} and Sun et al. \cite{sun2023research} also use CD as the loss function to measure the difference between point clouds to be registered. Based on the introduced deep spatial correlation representation feature, Deep-3DAligner \cite{wang2020unsupervised} describes the geometric essence of the spatial correlation between the source and the target point clouds in a coding-free manner. Sun et al. \cite{sun2023research} further develop the PointNetLK algorithm for cross-source DL-PCR. Moreover, to alleviate the problem of feature redundancy in DL-PCR, MMMI \cite{yuan2024MMIPCR} designs a structure aimed at maximizing the multi-level mutual information between features at different levels. Instead of CD, PCRNet \cite{sarode2019pcrnet} uses Earth Mover's Distance as the loss function, aiming to reduce the difference in features obtained through the PointNet \cite{qi2017pointnet}. 

\subsection{Correspondence-Based}

Correspondence-based unsupervised methods begin by extracting features, followed by the establishment of correspondences at various levels—point-level, distribution-level, or cluster-level. Rigid transformation parameters are then estimated based on these correspondences. These methods can be further categorized into RGB-D methods, probability model methods, descriptor-based methods, and geometry consistency-based methods.

\subsubsection{RGB-D} 
Motivated by multi-view geometry, the inherent geometric and photometric consistency in RGB-D sequences provides an effective self-supervised signal for point cloud feature extraction, which is therefore leveraged in unsupervised DL-PCR algorithms \cite{chen2024rgbd, el2021bootstrap}. To the best of our knowledge, UnsupervisedR\&R \cite{el2021unsupervisedr} is the first unsupervised algorithm to address RGB-D DL-PCR. It generates projected images through differentiable rendering and optimizes the registration network by backpropagating the rendering error. LLT \cite{wang2022improving} progressively fuses the learned geometric information from the point cloud and the visual information from the RGB-D data using a multi-scale local linear transformation. The resultant visual-geometric features exhibit reduced visual discrepancies caused by geometric changes in a canonical feature space, thus facilitating more reliable correspondences. LLT primarily focuses on using depth information to guide RGB processing but overlooks the interaction between the two modalities \cite{yuan2023pointmbf}. With that in mind, PointMBF \cite{yuan2023pointmbf} introduces a complex fusion strategy that integrates multi-scale and multi-directional features. On the other hand, Yan et al. \cite{yan2024discriminative} estimate point cloud correspondences by using hyperrectangle-based embeddings and intersection volume calculation, the complementary information from the RGB-D modalities are leveraged. As opposed to the aforementioned frame-to-frame consistency strategy, NeRF-UR \cite{yu2024nerf} leverages the neural radiance field (NeRF) \cite{mildenhall2021nerf} as a global model of the scene. It performs pose optimization by ensuring consistency between the input frames and the frames re-rendered by NeRF.

\subsubsection{Probability Model} 
\begin{figure}
	\centering
	\includegraphics[scale=0.745]{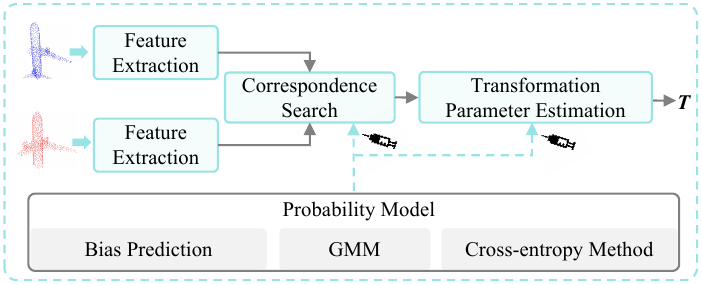}
	\caption{Visualization of the correspondence-based unsupervised registration process using a probabilistic model for optimization.}
	\label{Fig:cbu}
\end{figure}

As depicted in Fig. \ref{Fig:cbu}, probability model-based algorithms integrate probabilistic knowledge within the registration framework. These approaches utilize probability models to portray the matching relationship and invariant uncertainty between the point clouds. OBMNet \cite{shi2023overlap} proposes the overlap bias matching module, which first generates class probabilities and temperature parameters. Then, the Gumbel-Softmax \cite{jang2016categorical} is used to infer the distribution of shared structures. UGMM \cite{huang2022unsupervised} presents a novel approach, redefining the DL-PCR challenge as a clustering problem. Unlike UGMM which treats all clusters equally, UDPReg \cite{mei2023unsupervised} aligns different clusters with different importance. Additionally, self-consistency loss and cross-consistency loss are utilized to encourage the extracted features to be geometrically consistent. LatentCEM \cite{jiang2021planning} and CEMNet \cite{jiang2021sampling} model DL-PCR as a Markov decision process and use a cross-entropy method to find the optimal transformation iteratively.

\subsubsection{Descriptor-Based} 
Descriptor-based algorithms primarily focus on extracting robust feature descriptors from point clouds and leveraging these descriptors to establish accurate correspondences. The essence of these algorithms lies in the selection and matching strategies of features, which aim to achieve high-quality registration through accurate feature matching. DeepUME \cite{lang2021deepume} combines deep learning and universal manifold embedding \cite{efraim2019universal} registration methods to optimize feature consistency and enhance the robustness of registration. Furthermore, GTINet \cite{jiang2024gtinet} captures global topological relationships to guide the correspondence between the source and target point clouds. In CorrNet3D \cite{zeng2021corrnet3d}, the learned dense correspondences and feature consistency are utilized to drive unsupervised 3D registration through deformation-like reconstruction without the need for annotated data. R-PointHop \cite{kadam2022r} utilizes a local reference frame to achieve rotation and translation invariance. By extracting hierarchical features through point downsampling strategy, neighborhood expansion, and dimensionality reduction, R-PointHop builds robust point correspondences.

\subsubsection{Geometric Consistency-Based} Geometric consistency-based algorithms emphasize the preservation and exploitation of geometric consistency across local and global structures. By exploiting geometric differences between the source and pseudo-target point cloud neighborhoods, RIENet~\cite{shen2022reliable} designs an reliable inlier evaluation module and optimizes the model with global alignment, neighborhood consensus, and spatial consistency losses. In ICC \cite{yuan2024inlier}, the transformation-invariant geometric constraints and geometric structure consistency are constructed, to alleviate the inlier misinterpretation issues caused by relying on coordinate differences. By using a geometry-based local-to-global transformer, RegiFormer \cite{zheng2024regiformer} aggregates local neighborhood features and extracts global interrelationships within the entire point cloud. EYOC \cite{liu2024extend} utilizes structural information of the point cloud by progressively self-supervising distance extension, which employs spatial filtering and nearest neighbor search to enhance the accuracy and generalization of distant registration. In comparison, INTEGER \cite{xiong2024mining} further strengthens geometric consistency by integrating both high-level feature representations and low-level geometric cues, facilitating the reliable mining of pseudo-labels to enhance the inlier identification process.

\begin{table*}
	\centering
	\caption{Results on 3DMatch and 3DLoMatch.}
    \renewcommand{\arraystretch}{1}
	\begin{tabular}{l|cccccc|cccccc|c}
		\toprule
        \multirow{2}*{Method}&\multicolumn{6}{c|}{3DMatch}&\multicolumn{6}{c|}{3DLoMatch}&\multirow{2}*{Param.(M) $\downarrow$}\\
       % \cline{2-13} 
        \cmidrule{2-13}
		&5000&2500&1000&500&250&Average&5000&2500&1000&500&250&Average&\\
        \cmidrule{1-14}
        \multicolumn{14}{c}{Registration Recall (\%) $\uparrow$}\\
        \cmidrule{1-14}
        FCGF \cite{choy2019fully}&85.03&84.87&83.15&81.60&71.34&81.20&40.28&40.98&38.14&35.39&26.91&36.34&8.76\\
        Predator \cite{huang2021predator} &89.52&89.91&88.98&88.51&85.46&88.48&60.78&60.83&58.00&55.31&50.19&57.02&7.43\\
        CoFiNet \cite{yu2021cofinet}&89.33&88.96&88.43&87.35&87.12&88.24&67.33&66.21&64.16&63.08&61.04&64.36&5.48\\
        Lepard \cite{li2022lepard}&-&-&-&-&-&{93.43}&-&-&-&-&-&70.88&37.55\\
        REGTR \cite{yew2022regtr}&-&-&-&-&-&91.24&-&-&-&-&-&62.71&11.81\\
        GeoTransformer \cite{qin2023geotransformer}&91.71&91.45&91.16&91.11&91.17&91.32&72.95&73.51&72.79&73.74&72.46&{73.09}&9.83\\
        YOHO \cite{wang2022you}&90.77&90.32&89.24&88.52&84.53&88.68&65.04&65.61&63.16&56.55&48.12&59.70&12.38\\
        RoReg \cite{wang2023roreg} &92.94&93.10&92.93&93.18&91.15&92.66&70.54&70.70&68.31&67.95&64.90&68.48&12.71\\
        RoITr \cite{yu2023rotation}&91.82&91.55&91.37&90.86&91.23&91.37&72.25&72.84&72.51&72.09&71.70&72.28&10.10\\
        BUFFER \cite{ao2023buffer}&-&-&-&-&-&91.79&-&-&-&-&-&71.92&{0.92}\\
        SIRA-PCR \cite{chen2023sira}&93.68&93.68&93.11&93.05&92.61&93.23&74.42&75.09&75.12&73.92&71.76&74.06&-\\
        HEPCG \cite{xie2024hecpg}&90.65&90.18&90.73&90.18&90.00&90.35&72.20&72.47&72.44&72.04&71.62&72.15&16.59\\
        CAST \cite{huang2024consistency}&92.46&91.89&92.22&92.97&92.86&92.48&71.90&71.85&72.10&71.69&72.22&71.95&8.61\\
        Diff-Reg \cite{wu2024diff} &-&-&-&-&-&{94.29}&-&-&-&-&-&{73.22}&44.87\\
        PARE-Net \cite{yao2025pare}&-&-&-&-&-&{94.17}&-&-&-&-&-&{73.65}&{3.84}\\
        PointRegGPT \cite{chen2024pointreggpt}&92.44&92.11&91.86&91.39&90.16&91.59&76.76&76.19& 76.95&75.28&73.93&75.82&-\\
        \cmidrule{1-14}
        \multicolumn{14}{c}{Inlier Ratio (\%) $\uparrow$}\\
        \cmidrule{1-14}
        FCGF \cite{choy2019fully}&56.88&54.07&48.71&42.53&34.09&47.26&20.94&20.06&17.24&14.91&11.63&16.96&8.76\\
        Predator \cite{huang2021predator} &57.59&59.31&58.33&55.71&51.46&56.48&22.26&23.79&24.34&23.84&22.38&23.32&7.43\\
        CoFiNet \cite{yu2021cofinet}&49.85&50.99&51.73&51.90&52.11&51.32&24.04&25.25&25.97&26.19&26.17&25.52&5.48\\
        Lepard \cite{li2022lepard}&-&-&-&-&-&57.62&-&-&-&-&-&27.84&37.55\\
        GeoTransformer \cite{qin2023geotransformer}&72.62&79.14&83.81&85.62&86.82&81.60&43.83&49.43&55.80&58.58&60.54&53.64&9.83\\
        YOHO \cite{wang2022you}&64.40&60.73&55.82&47.33&42.77&54.21&26.04&23.12&22.64&18.64&16.17&21.32&12.38\\
        RoReg \cite{wang2023roreg} &81.58&80.68&75.14&74.50&75.92&77.64&39.64&39.84&33.68&31.99&34.43&35.92&12.71\\
        RoITr \cite{yu2023rotation}&82.22&82.30&82.43&82.49&82.39&81.37&53.25&53.58&53.78&53.81&53.95&53.67&10.10\\
        SIRA-PCR \cite{chen2023sira}&71.41&78.61&83.67&85.67&87.01&81.27&43.57&49.70&56.72&58.98&60.98&53.99&-\\
        HEPCG \cite{xie2024hecpg} &72.43&78.74&83.31&85.12&86.29&81.18&43.59&48.93&55.19&57.82&59.70&53.05&16.59\\        
        CAST \cite{huang2024consistency}&90.14&90.14&90.15&90.51&92.24&{90.64}&65.07&65.07&65.07&65.08&65.52&{65.16}&8.61\\
        Diff-Reg \cite{wu2024diff} &-&-&-&-&-&25.72&-&-&-&-&-&9.29&44.87\\
        PARE-Net \cite{yao2025pare}&-&-&-&-&-&74.73&-&-&-&-&-&46.28&{3.84}\\
        PointRegGPT \cite{chen2024pointreggpt}&73.84&80.48&84.94&86.67&87.82&82.75&45.85&52.01 &58.18&60.81&62.55&55.88&-\\
        \cmidrule{1-14}
        \multicolumn{14}{c}{Feature Matching Recall (\%) $\uparrow$}\\
        \cmidrule{1-14}
        FCGF \cite{choy2019fully}&97.34&97.16&97.08&96.73&96.59&96.98&76.56&75.38&74.17&71.87&67.22&73.04&8.76\\
        Predator \cite{huang2021predator}&96.32&96.43&96.60&96.94&96.18&96.49&73.94&75.24&74.53&75.21&75.35&74.85&7.43\\
        CoFiNet \cite{yu2021cofinet}&98.13&98.26&98.05&98.13&98.34&98.18&83.11&83.52&83.28&83.04&82.55&83.10&5.48\\
        Lepard \cite{li2022lepard}&-&-&-&-&-&98.09&-&-&-&-&-&84.39&37.55\\
        GeoTransformer \cite{qin2023geotransformer}&98.21&97.89&97.89&97.89&97.91&97.96&88.43&88.50&88.61&88.29&88.37&{88.44}&9.83\\
        YOHO \cite{wang2022you}&98.17&97.48&97.52&97.81&96.11&97.42&79.37&78.19&76.33&73.91&70.04&75.57&12.38\\
        RoReg \cite{wang2023roreg} &98.24&98.04&98.33&98.18&97.81&98.12&82.15&82.01&81.15&81.27&80.54&81.42&12.71\\
        RoITr \cite{yu2023rotation}&97.88&97.82&97.83&97.77&97.53&97.77&88.39&88.13&88.42&88.04&88.37&88.27&10.10\\
        SIRA-PCR \cite{chen2023sira}&98.44&98.32&98.43&98.47&98.36&98.40&88.95&89.18&88.66&88.64&88.65&88.82&-\\
        HEPCG \cite{xie2024hecpg} &98.36&98.21&98.31&98.32&98.37&{98.31}&86.78&86.88&87.92&87.62&87.45&87.33&16.59\\
        CAST \cite{huang2024consistency}&97.78&97.78&97.78&97.72&97.54&97.72&81.36&81.36&81.81&84.11&84.33&82.59&8.61\\
        Diff-Reg \cite{wu2024diff} &-&-&-&-&-&95.38&-&-&-&-&-&68.66&44.87\\
        PARE-Net \cite{yao2025pare}&-&-&-&-&-&98.20&-&-&-&-&-&87.93&{3.84}\\
        PointRegGPT \cite{chen2024pointreggpt}&98.67&98.74&98.55&98.45&98.24&98.53&89.45&89.65 &89.54&89.37&88.72&89.35&-\\
        \bottomrule
	\end{tabular}
	\centering 
    \label{table:compare_3dmatch}
\end{table*}

\section{Performance Comparison} \label{Sec-5}

To offer a clear comparison and fair evaluation of existing algorithms, we assess recent state-of-the-art DL-PCR methods through unified experimental setups, utilizing widely recognized benchmark datasets.

\subsection{Experimental Settings}

For the 3DMatch dataset, we also include its low-overlap variant, 3DLoMatch \cite{huang2021predator}. The preprocessing procedures follow those outlined in \cite{huang2021predator}. To ensure a fair comparison, we perform all experiments in a unified environment with a batch size of 1 and 40 epochs during training. Other experimental settings adhere to the default configurations specified in the respective algorithm code bases. The Python version used is 3.8, and the PyTorch version is 1.13.0. All experiments are conducted on a single NVIDIA GeForce RTX 4090 GPU. In the 3DMatch dataset, point cloud pairs exhibit an overlap greater than 30\%, while in 3DLoMatch, the overlap ranges from 10\% to 30\%. To evaluate performance, we use three metrics: 1) \textbf{Inlier Ratio}: the proportion of putative correspondences with residuals below a given threshold (e.g., 0.1m); 2) \textbf{Feature Matching Recall}: the percentage of point cloud pairs with an Inlier Ratio exceeding a specified threshold (e.g., 5\%); and 3) \textbf{Registration Recall}: the fraction of point cloud pairs for which the transformation error is below a predefined threshold (e.g., Root Mean Squared Error $<$ 0.2m). For algorithms that use RANSAC, results are reported for different numbers of correspondences (5000, 2500, 1000, 500, and 250). Other algorithms report results using uniform values. Additionally, the pre-trained methods SIRA-PCR \cite{chen2023sira} and PointRegGPT \cite{chen2024pointreggpt} are executed on GeoTransformer \cite{qin2023geotransformer}.

For the KITTI dataset, most algorithms follow the same experimental setup and directly reuse the results presented by comparison algorithms. To facilitate an intuitive comparison, we report the best metric from either the results in the original papers of each algorithm or those reproduced in subsequent works. In the experimental setup for the KITTI dataset, scene sequences from 11 cities are utilized, with sequences 0-5, 6-7, and 8-10 designated for training, validation, and testing, respectively. Three evaluation metrics are provided: \textbf{Relative Rotation Error (RRE)}, \textbf{Relative Translation Error (RTE)}, and \textbf{Registration Recall (RR)}. The RR is computed using a rotation threshold of 5$^\circ$ and a translation threshold of 2m.

\subsection{Results on 3DMatch and 3DLoMatch} 

Table \ref{table:compare_3dmatch} shows the experimental results for the 3DMatch and 3DLoMatch datasets. As shown, Diff-Reg, Lepard, and PARE-Net achieve high Registration Recall on 3DMatch. Diff-Reg and Lepard excel in handling variable DL-PCR tasks, but their performance is accompanied by a large number of parameters. On the other hand, PARE-Net achieves the best overall performance on both 3DMatch and 3DLoMatch, though its Inlier Ratio lags behind CAST, suggesting potential for improvement in correspondence precision. Regarding Feature Matching Recall, most algorithms perform exceptionally well, with results exceeding 95\% across the board, showcasing their robustness in identifying correct correspondences. However, no algorithm has optimized all metrics simultaneously. This indicates a need for a balanced approach that excels across key metrics while maintaining efficiency on both the 3DMatch and 3DLoMatch.

\subsection{Results on KITTI} 
Table \ref{table:compare_kitti} shows the experimental results comparing different algorithms on the KITTI dataset. As shown in the table, most algorithms achieve a 99.8\% Registration Recall (RR). This indicates that these algorithms can effectively match the majority of the point cloud data during registration, meeting the criteria of rotation errors below 5$^\circ$ and translation errors below 2m. However, there are noticeable differences in the Relative Rotation Error (RRE) and Relative Translation Error (RTE) across the algorithms. For instance, both OIF-PCR and PARE-Net excel in minimizing rotation and translation errors.  Notably, DCATr achieves the best performance in rotation error, with a value of 0.22$^\circ$, while also maintaining relatively low translation error.

\begin{table}
	\centering
	\caption{Results on KITTI.}
\renewcommand{\arraystretch}{1}
	\begin{tabular}{l|ccc}
		\toprule
        Method&RRE ($^{\circ}$) $\downarrow$&RTE (cm) $\downarrow$&RR (\%) $\uparrow$\\
       \midrule
        FCGF \cite{choy2019fully}&0.30&9.5&96.6\\
        D3Feat \cite{bai2020d3feat}&0.30&7.2&{99.8}\\
        DGR \cite{choy2020deep}&0.37&32.0&98.7\\
        PCAM \cite{cao2021pcam}&0.79&12.0&98.0\\
        HRegNet \cite{lu2021hregnet} &0.29&12.0&99.7\\
        CoFiNet \cite{yu2021cofinet}&0.41&8.2&{99.8}\\
        SpinNet \cite{ao2021spinnet}&0.47&9.9&99.1\\
        Predator \cite{huang2021predator} &0.27&6.8&{99.8}\\
        PointDSC \cite{bai2021pointdsc}&0.35&8.1&98.2\\
        DIP \cite{poiesi2021distinctive} &0.44&8.7&97.3\\
        GeDi \cite{poiesi2022learning} &0.32&7.2&99.8\\
        GeoTransformer \cite{qin2023geotransformer} &0.24&6.8&{99.8}\\
        OIF-PCR \cite{yang2022one} &0.23&6.5&{99.8}\\
        SC$^2$-PCR++ \cite{chen2023sc}&0.32&7.2&99.6\\
        UDPReg \cite{mei2023unsupervised}&0.41&8.8&64.6\\
        RegFormer \cite{liu2023regformer}&0.24&8.4&{99.8}\\
        VBReg \cite{jiang2023robust}&0.32&7.2&98.9\\
        DiffusionPCR \cite{chen2023diffusionpcr}&0.23&6.3&99.8\\
        PEAL \cite{yu2023peal}&0.23&6.8&99.8\\
        RIGA \cite{yu2024riga} &0.45&13.5&99.1\\
        RoCNet++ \cite{slimani2024rocnet++}&0.23&7.3&99.8\\
        MAC \cite{yang2024mac}&0.40&8.3&99.3\\
        DCATr\cite{chen2024dynamic}&{0.22}&6.6&99.7\\
        FastMAC\cite{zhang2024fastmac}&0.41&8.6&98.0\\%再看，里面的数据和别人的不同
        TEAR \cite{huang2024scalable}&0.39&8.6&99.1\\
        SPEAL \cite{xiong2024speal} &0.23&6.9&99.8\\
        PosDiffNet \cite{she2024posdiffnet}&0.24&6.6&{99.8}\\
        PARE-Net \cite{yao2025pare}&0.23&{4.9}&{99.8}\\
        \bottomrule
	\end{tabular}	
	\centering
    \label{table:compare_kitti}
\end{table}

\section{Challenges and Opportunities} \label{Sec-6}

While existing DL-PCR algorithms have yielded impressive results, several challenges remain. In this section, we highlight these challenges and outline potential research directions that could drive future advancements in the field.

\textbf{Data Generation.} Current data generation techniques for DL-PCR primarily rely on synthetic rendering and generative models to produce large-scale paired datasets \cite{chen2023sira, chen2024pointreggpt}, thus avoiding the high costs associated with real-world data collection. However, methods based on generative adversarial networks (GANs) and diffusion models face several challenges, including unstable training, slow data generation, and discrepancies between synthetic and real-world data distributions. Additionally, these approaches struggle to generate multi-view point cloud groups, limiting their applicability in complex and dynamic scenarios. To address these limitations, future research could focus on: (i) employing hybrid strategies that combine multiple methods to enhance realism and diversity, (ii) integrating physical simulations to accurately replicate sensor characteristics, (iii) incorporating geometric priors into real-world data to ensure consistency, and (iv) developing frameworks for multi-view consistency and dynamic scene generation to better support multi-view DL-PCR tasks.

\textbf{Multimodal Information.} Current multimodal DL-PCR algorithms improve feature representation by integrating image textures, which leads to more accurate and detailed mappings. Future research could focus on expanding the registration framework by incorporating additional modalities. For instance, (i) topological information from meshes can provide valuable structural insights, while (ii) semantic information derived from object recognition or segmentation can offer contextual understanding, aiding the model in distinguishing between relevant and irrelevant features. By integrating these modalities, the model can access richer contextual and structural information, thereby enhancing its ability to accurately estimate transformation parameters. 

\textbf{Vision-Language Models.} Large-scale vision-language models, such as CLIP \cite{radford2021learning}, establish semantic associations between images and text, enabling a deeper understanding of high-level semantic features. Although these models have shown promising results in point cloud recognition \cite{zhang2022pointclip, huang2023clip2point, zeng2023clip2}, their application to DL-PCR tasks remains largely unexplored. Future research could focus on integrating vision-language models into DL-PCR to enhance feature matching and scene understanding by providing global semantic information, particularly in complex or sparse data scenarios. 

\textbf{Semi-Supervised/Unsupervised Learning.} To enhance the scalability and applicability of DL-PCR models, it is essential to reduce the reliance on labeled datasets. As shown in our taxonomy in Figs. \ref{fig:spcr} and \ref{fig:unpcr}, supervised methods currently outnumber unsupervised approaches. To address this imbalance, future research should focus on developing semi-supervised and unsupervised learning techniques that can effectively learn from unlabeled or partially labeled data. Techniques such as self-supervised learning, unsupervised domain adaptation, and data augmentation offer promising potential for improving model performance, especially in situations where labeled data is limited or unavailable.

%\textbf{Pre-trained Models.} 

\section{Conclusion} \label{Sec-7}

This paper provides a comprehensive review of deep learning-based point cloud registration (DL-PCR) algorithms, covering over 100 registration methods and evaluating their performance across several commonly used benchmark datasets. It also highlights key challenges and outlines future research directions. To systematically organize the field and connect existing work, we propose a fine-grained taxonomy that categorizes algorithms based on their primary contributions and offers an analysis of their core ideas. Additionally, we provide an open-source repository with useful links to DL-PCR algorithms and datasets. We hope this survey will inspire and drive further research and innovation in this field.

\bibliographystyle{IEEEtran}

\bibliography{citepaper} 

\begin{thebibliography}{100}
\providecommand{\url}[1]{#1}
\csname url@samestyle\endcsname
\providecommand{\newblock}{\relax}
\providecommand{\bibinfo}[2]{#2}
\providecommand{\BIBentrySTDinterwordspacing}{\spaceskip=0pt\relax}
\providecommand{\BIBentryALTinterwordstretchfactor}{4}
\providecommand{\BIBentryALTinterwordspacing}{\spaceskip=\fontdimen2\font plus
\BIBentryALTinterwordstretchfactor\fontdimen3\font minus \fontdimen4\font\relax}
\providecommand{\BIBforeignlanguage}[2]{{%
\expandafter\ifx\csname l@#1\endcsname\relax
\typeout{** WARNING: IEEEtran.bst: No hyphenation pattern has been}%
\typeout{** loaded for the language `#1'. Using the pattern for}%
\typeout{** the default language instead.}%
\else
\language=\csname l@#1\endcsname
\fi
#2}}
\providecommand{\BIBdecl}{\relax}
\BIBdecl

\bibitem{guo2020deep}
Y.~Guo, H.~Wang, Q.~Hu, H.~Liu, L.~Liu, and M.~Bennamoun, ``Deep learning for 3d point clouds: A survey,'' \emph{IEEE Trans. Pattern Anal. Mach. Intell.}, vol.~43, no.~12, pp. 4338--4364, 2020.

\bibitem{xiao2023unsupervised}
A.~Xiao, J.~Huang, D.~Guan, X.~Zhang, S.~Lu, and L.~Shao, ``Unsupervised point cloud representation learning with deep neural networks: A survey,'' \emph{IEEE Trans. Pattern Anal. Mach. Intell.}, vol.~45, no.~9, pp. 11\,321--11\,339, 2023.

\bibitem{besl1992method}
P.~Besl and N.~McKay, ``Method for registration of 3-d shapes,'' \emph{IEEE Trans. Pattern Anal. Mach. Intell.}, vol.~14, no.~2, 1992.

\bibitem{jian2010robust}
B.~Jian and B.~C. Vemuri, ``Robust point set registration using gaussian mixture models,'' \emph{IEEE Trans. Pattern Anal. Mach. Intell.}, vol.~33, no.~8, pp. 1633--1645, 2010.

\bibitem{chen20203d}
S.~Chen, B.~Liu, C.~Feng, C.~Vallespi-Gonzalez, and C.~Wellington, ``3d point cloud processing and learning for autonomous driving,'' \emph{arXiv preprint arXiv:2003.00601}, 2020.

\bibitem{huang2024surface}
Z.~Huang, Y.~Wen, Z.~Wang, J.~Ren, and K.~Jia, ``Surface reconstruction from point clouds: A survey and a benchmark,'' \emph{IEEE Trans. Pattern Anal. Mach. Intell.}, vol.~46, no.~12, pp. 9727--9748, 2024.

\bibitem{dang2022learning}
Z.~Dang, L.~Wang, Y.~Guo, and M.~Salzmann, ``Learning-based point cloud registration for 6d object pose estimation in the real world,'' in \emph{Proc. Eur. Conf. Comput. Vis.}\hskip 1em plus 0.5em minus 0.4em\relax Springer, 2022, pp. 19--37.

\bibitem{elbaz20173d}
G.~Elbaz, T.~Avraham, and A.~Fischer, ``3d point cloud registration for localization using a deep neural network auto-encoder,'' in \emph{Proc. IEEE Conf. Comput. Vis. Pattern Recognit.}, 2017, pp. 4631--4640.

\bibitem{gomes20143d}
L.~Gomes, O.~R.~P. Bellon, and L.~Silva, ``3d reconstruction methods for digital preservation of cultural heritage: A survey,'' \emph{Pattern Recognit. Lett.}, vol.~50, pp. 3--14, 2014.

\bibitem{pomerleau2015review}
F.~Pomerleau, F.~Colas, R.~Siegwart \emph{et~al.}, ``A review of point cloud registration algorithms for mobile robotics,'' \emph{Found. Trends Robot.}, vol.~4, no.~1, pp. 1--104, 2015.

\bibitem{sinko20183d}
M.~Sinko, P.~Kamencay, R.~Hudec, and M.~Benco, ``3d registration of the point cloud data using icp algorithm in medical image analysis,'' in \emph{2018 ELEKTRO}.\hskip 1em plus 0.5em minus 0.4em\relax IEEE, 2018, pp. 1--6.

\bibitem{lowens2024unsupervised}
C.~L{\"o}wens, T.~Funke, A.~Wagner, and A.~P. Condurache, ``Unsupervised point cloud registration with self-distillation,'' \emph{arXiv preprint arXiv:2409.07558}, 2024.

\bibitem{ao2023buffer}
S.~Ao, Q.~Hu, H.~Wang, K.~Xu, and Y.~Guo, ``Buffer: Balancing accuracy, efficiency, and generalizability in point cloud registration,'' in \emph{Proc. IEEE Conf. Comput. Vis. Pattern Recognit.}, 2023, pp. 1255--1264.

\bibitem{qin2023geotransformer}
Z.~Qin, H.~Yu, C.~Wang, Y.~Guo, Y.~Peng, S.~Ilic, D.~Hu, and K.~Xu, ``Geotransformer: Fast and robust point cloud registration with geometric transformer,'' \emph{IEEE Trans. Pattern Anal. Mach. Intell.}, vol.~45, no.~8, pp. 9806--9821, 2023.

\bibitem{zhang2024constructing}
Y.-X. Zhang, J.~Gui, and J.~T.-Y. Kwok, ``Constructing diverse inlier consistency for partial point cloud registration,'' \emph{IEEE Trans. Image Process.}, vol.~33, pp. 6535 -- 6549, 2024.

\bibitem{chen2023full}
G.~Chen, M.~Wang, Q.~Zhang, L.~Yuan, and Y.~Yue, ``Full transformer framework for robust point cloud registration with deep information interaction,'' \emph{IEEE Trans. Neural Netw. Learn. Syst.}, vol.~35, no.~10, pp. 13\,368--13\,382, 2023.

\bibitem{pomerleau2012challenging}
F.~Pomerleau, M.~Liu, F.~Colas, and R.~Siegwart, ``Challenging data sets for point cloud registration algorithms,'' \emph{Int. J. Robot. Res.}, vol.~31, no.~14, pp. 1705--1711, 2012.

\bibitem{geiger2012we}
A.~Geiger, P.~Lenz, and R.~Urtasun, ``Are we ready for autonomous driving? the kitti vision benchmark suite,'' in \emph{Proc. IEEE Conf. Comput. Vis. Pattern Recognit.}\hskip 1em plus 0.5em minus 0.4em\relax IEEE, 2012, pp. 3354--3361.

\bibitem{choi2015robust}
S.~Choi, Q.-Y. Zhou, and V.~Koltun, ``Robust reconstruction of indoor scenes,'' in \emph{Proc. IEEE Conf. Comput. Vis. Pattern Recognit.}, 2015, pp. 5556--5565.

\bibitem{wu20153d}
Z.~Wu, S.~Song, A.~Khosla, F.~Yu, L.~Zhang, X.~Tang, and J.~Xiao, ``3d shapenets: A deep representation for volumetric shapes,'' in \emph{Proc. IEEE Conf. Comput. Vis. Pattern Recognit.}, 2015, pp. 1912--1920.

\bibitem{chang2015shapenet}
A.~Chang, T.~Funkhouser, L.~Guibas, P.~Hanrahan, Q.~Huang, Z.~Li, S.~Savarese, M.~Savva, S.~Song, H.~Su, X.~J, Y.~L., and Y.~F., ``Shapenet: An information-rich 3d model repository,'' \emph{arXiv preprint arXiv:1512.03012}, 2015.

\bibitem{choi2016large}
S.~Choi, Q.-Y. Zhou, S.~Miller, and V.~Koltun, ``A large dataset of object scans,'' \emph{arXiv preprint arXiv:1602.02481}, 2016.

\bibitem{zeng20173dmatch}
A.~Zeng, S.~Song, M.~Nie{\ss}ner, M.~Fisher, J.~Xiao, and T.~Funkhouser, ``3dmatch: Learning local geometric descriptors from rgb-d reconstructions,'' in \emph{Proc. IEEE Conf. Comput. Vis. Pattern Recognit.}, 2017, pp. 1802--1811.

\bibitem{maddern20171}
W.~Maddern, G.~Pascoe, C.~Linegar, and P.~Newman, ``1 year, 1000 km: The oxford robotcar dataset,'' \emph{Int. J. Robot. Res.}, vol.~36, no.~1, pp. 3--15, 2017.

\bibitem{uy2019revisiting}
M.~A. Uy, Q.-H. Pham, B.-S. Hua, T.~Nguyen, and S.-K. Yeung, ``Revisiting point cloud classification: A new benchmark dataset and classification model on real-world data,'' in \emph{Proc. IEEE Int. Conf. Comput. Vis.}, 2019, pp. 1588--1597.

\bibitem{dong2020registration}
Z.~Dong, F.~Liang, B.~Yang, Y.~Xu, Y.~Zang, J.~Li, Y.~Wang, W.~Dai, H.~Fan, J.~Hyypp{\"a}, and S.~U., ``Registration of large-scale terrestrial laser scanner point clouds: A review and benchmark,'' \emph{ISPRS J. Photogramm. Remote Sens.}, vol. 163, pp. 327--342, 2020.

\bibitem{caesar2020nuscenes}
H.~Caesar, V.~Bankiti, A.~H. Lang, S.~Vora, V.~E. Liong, Q.~Xu, A.~Krishnan, Y.~Pan, G.~Baldan, and O.~Beijbom, ``Nuscenes: A multimodal dataset for autonomous driving,'' in \emph{Proc. IEEE Conf. Comput. Vis. Pattern Recognit.}, 2020, pp. 11\,621--11\,631.

\bibitem{pan2021variational}
L.~Pan, X.~Chen, Z.~Cai, J.~Zhang, H.~Zhao, S.~Yi, and Z.~Liu, ``Variational relational point completion network,'' in \emph{Proc. IEEE Conf. Comput. Vis. Pattern Recognit.}, 2021, pp. 8524--8533.

\bibitem{chen2023sira}
S.~Chen, H.~Xu, R.~Li, G.~Liu, C.-W. Fu, and S.~Liu, ``Sira-pcr: Sim-to-real adaptation for 3d point cloud registration,'' in \emph{Proc. IEEE Conf. Comput. Vis. Pattern Recognit.}, 2023, pp. 14\,394--14\,405.

\bibitem{gu2020review}
X.~Gu, X.~Wang, and Y.~Guo, ``A review of research on point cloud registration methods,'' in \emph{IOP Conf. Ser.: Mater. Sci. Eng.}, vol. 782, no.~2.\hskip 1em plus 0.5em minus 0.4em\relax IOP Publishing, 2020, p. 022070.

\bibitem{bellekens2014survey}
B.~Bellekens, V.~Spruyt, R.~Berkvens, and M.~Weyn, ``A survey of rigid 3d pointcloud registration algorithms,'' in \emph{Proc. Int. Conf. Ambient Comput., Appl., Serv. Technol.}, 2014, pp. 8--13.

\bibitem{tam2012registration}
G.~K. Tam, Z.-Q. Cheng, Y.-K. Lai, F.~C. Langbein, Y.~Liu, D.~Marshall, R.~R. Martin, X.-F. Sun, and P.~L. Rosin, ``Registration of 3d point clouds and meshes: A survey from rigid to nonrigid,'' \emph{IEEE Trans. Vis. Comput. Graph.}, vol.~19, no.~7, pp. 1199--1217, 2012.

\bibitem{huang2021comprehensive}
X.~Huang, G.~Mei, J.~Zhang, and R.~Abbas, ``A comprehensive survey on point cloud registration,'' \emph{arXiv preprint arXiv:2103.02690}, 2021.

\bibitem{lyu2024rigid}
M.~Lyu, J.~Yang, Z.~Qi, R.~Xu, and J.~Liu, ``Rigid pairwise 3d point cloud registration: a survey,'' \emph{Pattern Recognit.}, p. 110408, 2024.

\bibitem{zhao2024deep}
Y.~Zhao, J.~Zhang, S.~Xu, and J.~Ma, ``Deep learning-based low overlap point cloud registration for complex scenario: The review,'' \emph{Inf. Fusion}, vol. 107, p. 102305, 2024.

\bibitem{huang2023cross}
X.~Huang, G.~Mei, and J.~Zhang, ``Cross-source point cloud registration: Challenges, progress and prospects,'' \emph{Neurocomputing}, vol. 548, p. 126383, 2023.

\bibitem{zhang2020deep}
Z.~Zhang, Y.~Dai, and J.~Sun, ``Deep learning based point cloud registration: an overview,'' \emph{Virtual Real. Intell. Hardw.}, vol.~2, no.~3, pp. 222--246, 2020.

\bibitem{ijcai2024p922}
Y.-X. Zhang, J.~Gui, X.~Cong, X.~Gong, and W.~Tao, ``A comprehensive survey and taxonomy on point cloud registration based on deep learning,'' in \emph{Proc. Int. Joint Conf. Artif. Intell.}, 8 2024, pp. 8344--8353.

\bibitem{xiao2013sun3d}
J.~Xiao, A.~Owens, and A.~Torralba, ``Sun3d: A database of big spaces reconstructed using sfm and object labels,'' in \emph{Proc. IEEE Int. Conf. Comput. Vis.}, 2013, pp. 1625--1632.

\bibitem{shotton2013scene}
J.~Shotton, B.~Glocker, C.~Zach, S.~Izadi, A.~Criminisi, and A.~Fitzgibbon, ``Scene coordinate regression forests for camera relocalization in rgb-d images,'' in \emph{Proc. IEEE Conf. Comput. Vis. Pattern Recognit.}, 2013, pp. 2930--2937.

\bibitem{lai2014unsupervised}
K.~Lai, L.~Bo, and D.~Fox, ``Unsupervised feature learning for 3d scene labeling,'' in \emph{Proc. IEEE Int. Conf. Robot. Autom.}\hskip 1em plus 0.5em minus 0.4em\relax IEEE, 2014, pp. 3050--3057.

\bibitem{dai2017bundlefusion}
A.~Dai, M.~Nie{\ss}ner, M.~Zollh{\"o}fer, S.~Izadi, and C.~Theobalt, ``Bundlefusion: Real-time globally consistent 3d reconstruction using on-the-fly surface reintegration,'' \emph{ACM Trans. Graph}, vol.~36, no.~4, p.~1, 2017.

\bibitem{valentin2016learning}
J.~Valentin, A.~Dai, M.~Nie{\ss}ner, P.~Kohli, P.~Torr, S.~Izadi, and C.~Keskin, ``Learning to navigate the energy landscape,'' in \emph{Proc. Int. Conf.3D Vis.}\hskip 1em plus 0.5em minus 0.4em\relax IEEE, 2016, pp. 323--332.

\bibitem{sun2020circle}
Y.~Sun, C.~Cheng, Y.~Zhang, C.~Zhang, L.~Zheng, Z.~Wang, and Y.~Wei, ``Circle loss: A unified perspective of pair similarity optimization,'' in \emph{Proc. IEEE Conf. Comput. Vis. Pattern Recognit.}, 2020, pp. 6398--6407.

\bibitem{yu2023rotation}
H.~Yu, Z.~Qin, J.~Hou, M.~Saleh, D.~Li, B.~Busam, and S.~Ilic, ``Rotation-invariant transformer for point cloud matching,'' in \emph{Proc. IEEE Conf. Comput. Vis. Pattern Recognit.}, 2023, pp. 5384--5393.

\bibitem{xie2024hecpg}
Y.~Xie, J.~Zhu, S.~Li, N.~Hu, and P.~Shi, ``Hecpg: hyperbolic embedding and confident patch-guided network for point cloud matching,'' \emph{IEEE Trans. Geosci. Remote Sens.}, vol.~62, pp. 1--12, 2024.

\bibitem{huang2021predator}
S.~Huang, Z.~Gojcic, M.~Usvyatsov, A.~Wieser, and K.~Schindler, ``Predator: Registration of 3d point clouds with low overlap,'' in \emph{Proc. IEEE Int. Conf. Comput. Vis.}, 2021, pp. 4267--4276.

\bibitem{xu2021omnet}
H.~Xu, S.~Liu, G.~Wang, G.~Liu, and B.~Zeng, ``Omnet: Learning overlapping mask for partial-to-partial point cloud registration,'' in \emph{Proc. IEEE Int. Conf. Comput. Vis.}, 2021, pp. 3132--3141.

\bibitem{huang2024consistency}
R.~Huang, Y.~Tang, J.~Chen, and L.~Li, ``A consistency-aware spot-guided transformer for versatile and hierarchical point cloud registration,'' \emph{Proc. Conf. Neural Inf. Process. Syst.}, 2024.

\bibitem{yao2025pare}
R.~Yao, S.~Du, W.~Cui, C.~Tang, and C.~Yang, ``Pare-net: Position-aware rotation-equivariant networks for robust point cloud registration,'' in \emph{Proc. Eur. Conf. Comput. Vis.}\hskip 1em plus 0.5em minus 0.4em\relax Springer, 2025, pp. 287--303.

\bibitem{wang2019prnet}
Y.~Wang and J.~Solomon, ``Prnet: Self-supervised learning for partial-to-partial registration,'' in \emph{Proc. Int. Conf. Neural Inf. Process. Syst.}, vol.~32, 2019.

\bibitem{huang2022unsupervised}
X.~Huang, S.~Li, Y.~Zuo, Y.~Fang, J.~Zhang, and X.~Zhao, ``Unsupervised point cloud registration by learning unified gaussian mixture models,'' \emph{IEEE Robotics Autom. Lett.}, vol.~7, no.~3, pp. 7028--7035, 2022.

\bibitem{mei2023unsupervised}
G.~Mei, H.~Tang, X.~Huang, W.~Wang, J.~Liu, J.~Zhang, L.~Van~Gool, and Q.~Wu, ``Unsupervised deep probabilistic approach for partial point cloud registration,'' in \emph{Proc. IEEE Conf. Comput. Vis. Pattern Recognit.}, 2023, pp. 13\,611--13\,620.

\bibitem{sarode2019pcrnet}
V.~Sarode, X.~Li, H.~Goforth, Y.~Aoki, R.~Srivatsan, and S.~Lucey, ``Pcrnet: Point cloud registration network using pointnet encoding,'' \emph{arXiv preprint arXiv:1908.07906}, 2019.

\bibitem{el2021bootstrap}
M.~El~Banani and J.~Johnson, ``Bootstrap your own correspondences,'' in \emph{Proc. IEEE Int. Conf. Comput. Vis.}, 2021, pp. 6433--6442.

\bibitem{el2021unsupervisedr}
M.~El~Banani, L.~Gao, and J.~Johnson, ``Unsupervisedr\&r: Unsupervised point cloud registration via differentiable rendering,'' in \emph{Proc. IEEE Conf. Comput. Vis. Pattern Recognit.}, 2021, pp. 7129--7139.

\bibitem{deng2018ppfnet}
H.~Deng, T.~Birdal, and S.~Ilic, ``Ppfnet: Global context aware local features for robust 3d point matching,'' in \emph{Proc. IEEE Conf. Comput. Vis. Pattern Recognit.}, 2018, pp. 195--205.

\bibitem{gojcic2019perfect}
Z.~Gojcic, C.~Zhou, J.~Wegner, and A.~Wieser, ``The perfect match: 3d point cloud matching with smoothed densities,'' in \emph{Proc. IEEE Conf. Comput. Vis. Pattern Recognit.}, 2019, pp. 5545--5554.

\bibitem{bai2020d3feat}
X.~Bai, Z.~Luo, L.~Zhou, H.~Fu, L.~Quan, and C.-L. Tai, ``D3feat: Joint learning of dense detection and description of 3d local features,'' in \emph{Proc. IEEE Conf. Comput. Vis. Pattern Recognit.}, 2020, pp. 6359--6367.

\bibitem{ao2021spinnet}
S.~Ao, Q.~Hu, B.~Yang, A.~Markham, and Y.~Guo, ``Spinnet: Learning a general surface descriptor for 3d point cloud registration,'' in \emph{Proc. IEEE Conf. Comput. Vis. Pattern Recognit.}, 2021, pp. 11\,753--11\,762.

\bibitem{poiesi2021distinctive}
F.~Poiesi and D.~Boscaini, ``Distinctive 3d local deep descriptors,'' in \emph{Proc. Int. Conf. Pattern Recognit.}\hskip 1em plus 0.5em minus 0.4em\relax IEEE, 2021, pp. 5720--5727.

\bibitem{wang2022you}
H.~Wang, Y.~Liu, Z.~Dong, and W.~Wang, ``You only hypothesize once: Point cloud registration with rotation-equivariant descriptors,'' in \emph{Proc. ACM Int. Conf. Multimedia}, 2022, pp. 1630--1641.

\bibitem{poiesi2022learning}
F.~Poiesi and D.~Boscaini, ``Learning general and distinctive 3d local deep descriptors for point cloud registration,'' \emph{IEEE Trans. Pattern Anal. Mach. Intell.}, vol.~45, no.~3, pp. 3979--3985, 2022.

\bibitem{wang2023roreg}
H.~Wang, Y.~Liu, Q.~Hu, B.~Wang, J.~Chen, Z.~Dong, Y.~Guo, W.~Wang, and B.~Yang, ``Roreg: Pairwise point cloud registration with oriented descriptors and local rotations,'' \emph{IEEE Trans. Pattern Anal. Mach. Intell.}, vol.~45, no.~8, pp. 10\,376--10\,393, 2023.

\bibitem{zhao2023spherenet}
G.~Zhao, Z.~Guo, X.~Wang, and H.~Ma, ``Spherenet: Learning a noise-robust and general descriptor for point cloud registration,'' \emph{IEEE Trans. Geosci. Remote Sens.}, vol.~62, 2023.

\bibitem{pan2024robust}
L.~Pan, Z.~Cai, and Z.~Liu, ``Robust partial-to-partial point cloud registration in a full range,'' \emph{IEEE Robot. Autom. Lett.}, vol.~9, no.~3, pp. 2861--2868, 2024.

\bibitem{liu2024deep}
S.~Liu, T.~Wang, Y.~Zhang, R.~Zhou, L.~Li, C.~Dai, Y.~Zhang, L.~Wang, and H.~Wang, ``Deep semantic graph matching for large-scale outdoor point cloud registration,'' \emph{IEEE Trans. Geosci. Remote Sens.}, vol.~62, 2024.

\bibitem{slimani2024rocnet++}
K.~Slimani, C.~Achard, and B.~Tamadazte, ``Rocnet++: Triangle-based descriptor for accurate and robust point cloud registration,'' \emph{Pattern Recognit.}, vol. 147, p. 110108, 2024.

\bibitem{zhao2024ha}
B.~Zhao, Q.~Liu, Z.~Wang, X.~Chen, Z.~Jia, and D.~Liang, ``Ha-tinet: Learning a distinctive and general 3d local descriptor for point cloud registration,'' \emph{IEEE Trans. Vis. Comput. Graph.}, 2024.

\bibitem{wang2022storm}
Y.~Wang, C.~Yan, Y.~Feng, S.~Du, Q.~Dai, and Y.~Gao, ``Storm: Structure-based overlap matching for partial point cloud registration,'' \emph{IEEE Trans. Pattern Anal. Mach. Intell.}, vol.~45, no.~1, pp. 1135--1149, 2022.

\bibitem{wu2023rornet}
Y.~Wu, Y.~Zhang, W.~Ma, M.~Gong, X.~Fan, M.~Zhang, A.~Qin, and Q.~Miao, ``Rornet: Partial-to-partial registration network with reliable overlapping representations,'' \emph{IEEE Trans. Neural Netw. Learn. Syst.}, vol.~35, no.~11, pp. 15\,453--15\,466, 2023.

\bibitem{li2023unified}
L.~Li, W.~Ding, Y.~Wen, Y.~Liang, Y.~Liu, and G.~Wan, ``A unified bev model for joint learning of 3d local features and overlap estimation,'' in \emph{Proc. IEEE Int. Conf. Robot. Autom.}\hskip 1em plus 0.5em minus 0.4em\relax IEEE, 2023, pp. 8341--8348.

\bibitem{liu2024low}
Y.~Liu and Z.~Liu, ``Low overlapping point cloud registration using mutual prior based completion network,'' \emph{IEEE Trans. Image Process.}, vol.~33, pp. 4781--4795, 2024.

\bibitem{wu2021feature}
B.~Wu, J.~Ma, G.~Chen, and P.~An, ``Feature interactive representation for point cloud registration,'' in \emph{Proc. IEEE Int. Conf. Comput. Vis.}, 2021, pp. 5530--5539.

\bibitem{yang2022one}
F.~Yang, L.~Guo, Z.~Chen, and W.~Tao, ``One-inlier is first: Towards efficient position encoding for point cloud registration,'' \emph{Proc. Int. Conf. Neural Inf. Process. Syst.}, vol.~35, pp. 6982--6995, 2022.

\bibitem{zhang2022end}
Z.~Zhang, J.~Sun, Y.~Dai, D.~Zhou, X.~Song, and M.~He, ``End-to-end learning the partial permutation matrix for robust 3d point cloud registration,'' in \emph{Proc. AAAI Conf. Artif. Intell.}, vol.~36, no.~3, 2022, pp. 3399--3407.

\bibitem{lee2021deep}
J.~Lee, S.~Kim, M.~Cho, and J.~Park, ``Deep hough voting for robust global registration,'' in \emph{Proc. IEEE Int. Conf. Comput. Vis.}, 2021, pp. 15\,994--16\,003.

\bibitem{bai2021pointdsc}
X.~Bai, Z.~Luo, L.~Zhou, H.~Chen, L.~Li, Z.~Hu, H.~Fu, and C.-L. Tai, ``Pointdsc: Robust point cloud registration using deep spatial consistency,'' in \emph{Proc. IEEE Conf. Comput. Vis. Pattern Recognit.}, 2021, pp. 15\,859--15\,869.

\bibitem{zhang2022partial}
Y.~Zhang, Z.~Sun, Z.~Zeng, and K.~Lam, ``Partial point cloud registration with deep local feature,'' \emph{IEEE Trans. Artif. Intell.}, vol.~4, no.~5, pp. 1317--1327, 2022.

\bibitem{chen2022sc2}
Z.~Chen, K.~Sun, F.~Yang, and W.~Tao, ``Sc$^2$-pcr: A second order spatial compatibility for efficient and robust point cloud registration,'' in \emph{Proc. IEEE Conf. Comput. Vis. Pattern Recognit.}, 2022, pp. 13\,221--13\,231.

\bibitem{chen2023sc}
Z.~Chen, K.~Sun, F.~Yang, L.~Guo, and W.~Tao, ``Sc$^2$-pcr++: Rethinking the generation and selection for efficient and robust point cloud registration,'' \emph{IEEE Trans. Pattern Anal. Mach. Intell.}, vol.~45, no.~10, pp. 12\,358--12\,376, 2023.

\bibitem{yao2023hunter}
R.~Yao, S.~Du, W.~Cui, A.~Ye, F.~Wen, H.~Zhang, Z.~Tian, and Y.~Gao, ``Hunter: Exploring high-order consistency for point cloud registration with severe outliers,'' \emph{IEEE Trans. Pattern Anal. Mach. Intell.}, vol.~45, no.~12, pp. 14\,760--14\,776, 2023.

\bibitem{yuan2024robust}
M.~Yuan, K.~Fu, Z.~Li, Y.~Meng, A.~Shen, and M.~Wang, ``Robust point cloud registration via random network co-ensemble,'' \emph{IEEE Trans. Circuits Syst. Video Technol.}, 2024.

\bibitem{huang2024scalable}
T.~Huang, L.~Peng, R.~Vidal, and Y.-H. Liu, ``Scalable 3d registration via truncated entry-wise absolute residuals,'' in \emph{Proc. IEEE Conf. Comput. Vis. Pattern Recognit.}, 2024, pp. 27\,477--27\,487.

\bibitem{yang2024mac}
J.~Yang, X.~Zhang, P.~Wang, Y.~Guo, K.~Sun, Q.~Wu, S.~Zhang, and Y.~Zhang, ``Mac: Maximal cliques for 3d registration,'' \emph{IEEE Trans. Pattern Anal. Mach. Intell.}, vol.~46, no.~12, pp. 10\,645--10\,662, 2024.

\bibitem{zhang2024fastmac}
Y.~Zhang, H.~Zhao, H.~Li, and S.~Chen, ``Fastmac: Stochastic spectral sampling of correspondence graph,'' in \emph{Proc. IEEE Conf. Comput. Vis. Pattern Recognit.}, 2024, pp. 17\,857--17\,867.

\bibitem{chen2022detarnet}
Z.~Chen, F.~Yang, and W.~Tao, ``Detarnet: Decoupling translation and rotation by siamese network for point cloud registration,'' in \emph{Proc. AAAI Conf. Artif. Intell.}, vol.~36, no.~1, 2022, pp. 401--409.

\bibitem{xu2022finet}
H.~Xu, N.~Ye, G.~Liu, B.~Zeng, and S.~Liu, ``Finet: Dual branches feature interaction for partial-to-partial point cloud registration,'' in \emph{Proc. AAAI Conf. Artif. Intell.}, vol.~36, no.~3, 2022, pp. 2848--2856.

\bibitem{zhang2022self}
Z.~Zhang, J.~Sun, Y.~Dai, D.~Zhou, X.~Song, and M.~He, ``Self-supervised rigid transformation equivariance for accurate 3d point cloud registration,'' \emph{Pattern Recognit.}, vol. 130, p. 108784, 2022.

\bibitem{yuan2024learning}
Y.~Yuan, Y.~Wu, J.~Lei, C.~Hu, M.~Gong, X.~Fan, W.~Ma, and Q.~Miao, ``Learning compact transformation based on dual quaternion for point cloud registration,'' \emph{IEEE Trans. Instrum. Meas.}, vol.~73, 2024.

\bibitem{jin2024q}
S.~Jin, D.~Barath, M.~Pollefeys, and I.~Armeni, ``Q-reg: End-to-end trainable point cloud registration with surface curvature,'' in \emph{Proc. Int. Conf.3D Vis.}\hskip 1em plus 0.5em minus 0.4em\relax IEEE, 2024, pp. 1330--1339.

\bibitem{zhou2018learning}
L.~Zhou, S.~Zhu, Z.~Luo, T.~Shen, R.~Zhang, M.~Zhen, T.~Fang, and L.~Quan, ``Learning and matching multi-view descriptors for registration of point clouds,'' in \emph{Proc. Eur. Conf. Comput. Vis.}, 2018, pp. 505--522.

\bibitem{lu2019deepvcp}
W.~Lu, G.~Wan, Y.~Zhou, X.~Fu, P.~Yuan, and S.~Song, ``Deepvcp: An end-to-end deep neural network for point cloud registration,'' in \emph{Proc. IEEE Int. Conf. Comput. Vis.}, 2019, pp. 12--21.

\bibitem{gojcic2020learning}
Z.~Gojcic, C.~Zhou, J.~D. Wegner, L.~J. Guibas, and T.~Birdal, ``Learning multiview 3d point cloud registration,'' in \emph{Proc. IEEE Conf. Comput. Vis. Pattern Recognit.}, 2020, pp. 1759--1769.

\bibitem{lu2021hregnet}
F.~Lu, G.~Chen, Y.~Liu, L.~Zhang, S.~Qu, S.~Liu, and R.~Gu, ``Hregnet: A hierarchical network for large-scale outdoor lidar point cloud registration,'' in \emph{Proc. IEEE Int. Conf. Comput. Vis.}, 2021, pp. 16\,014--16\,023.

\bibitem{fu2021robust}
K.~Fu, S.~Liu, X.~Luo, and M.~Wang, ``Robust point cloud registration framework based on deep graph matching,'' in \emph{Proc. IEEE Conf. Comput. Vis. Pattern Recognit.}, 2021, pp. 8893--8902.

\bibitem{wang2023robust}
H.~Wang, Y.~Liu, Z.~Dong, Y.~Guo, Y.-S. Liu, W.~Wang, and B.~Yang, ``Robust multiview point cloud registration with reliable pose graph initialization and history reweighting,'' in \emph{Proc. IEEE Conf. Comput. Vis. Pattern Recognit.}, 2023, pp. 9506--9515.

\bibitem{yuan2020deepgmr}
W.~Yuan, B.~Eckart, K.~Kim, V.~Jampani, D.~Fox, and J.~Kautz, ``Deepgmr: Learning latent gaussian mixture models for registration,'' in \emph{Proc. Eur. Conf. Comput. Vis.}\hskip 1em plus 0.5em minus 0.4em\relax Springer, 2020, pp. 733--750.

\bibitem{mei2023overlap}
G.~Mei, F.~Poiesi, C.~Saltori, J.~Zhang, E.~Ricci, and N.~Sebe, ``Overlap-guided gaussian mixture models for point cloud registration,'' in \emph{Proc. IEEE/CVF Winter Conf. Appl. Comput. Vis.}, 2023, pp. 4511--4520.

\bibitem{chen2023point}
H.~Chen, B.~Chen, Z.~Zhao, and B.~Song, ``Point cloud registration based on learning gaussian mixture models with global-weighted local representations,'' \emph{IEEE Geosci. Remote Sens. Lett.}, vol.~20, pp. 1--5, 2023.

\bibitem{jiang2023robust}
H.~Jiang, Z.~Dang, Z.~Wei, J.~Xie, J.~Yang, and M.~Salzmann, ``Robust outlier rejection for 3d registration with variational bayes,'' in \emph{Proc. IEEE Conf. Comput. Vis. Pattern Recognit.}, 2023, pp. 1148--1157.

\bibitem{xu2023point}
J.~Xu, Y.~Zhang, Y.~Zou, and P.~X. Liu, ``Point cloud registration with zero overlap rate and negative overlap rate,'' \emph{IEEE Robot. Autom. Lett.}, 2023.

\bibitem{chen2023diffusionpcr}
Z.~Chen, Y.~Ren, T.~Zhang, Z.~Dang, W.~Tao, S.~S{\"u}sstrunk, and M.~Salzmann, ``Diffusionpcr: Diffusion models for robust multi-step point cloud registration,'' \emph{arXiv preprint arXiv:2312.03053}, 2023.

\bibitem{she2024posdiffnet}
R.~She, S.~Wang, Q.~Kang, K.~Zhao, Y.~Song, W.~P. Tay, T.~Geng, and X.~Jian, ``Posdiffnet: Positional neural diffusion for point cloud registration in a large field of view with perturbations,'' in \emph{Proc. AAAI Conf. Artif. Intell.}, vol.~38, no.~1, 2024, pp. 231--239.

\bibitem{she2024pointdifformer}
R.~She, Q.~Kang, S.~Wang, W.~P. Tay, K.~Zhao, Y.~Song, T.~Geng, Y.~Xu, D.~N. Navarro, and A.~Hartmannsgruber, ``Pointdifformer: Robust point cloud registration with neural diffusion and transformer,'' \emph{IEEE Trans. Geosci. Remote Sens.}, 2024.

\bibitem{jiang2024se}
H.~Jiang, M.~Salzmann, Z.~Dang, J.~Xie, and J.~Yang, ``Se (3) diffusion model-based point cloud registration for robust 6d object pose estimation,'' \emph{Proc. Int. Conf. Neural Inf. Process. Syst.}, vol.~36, 2024.

\bibitem{wu2024diff}
Q.~Wu, H.~Jiang, L.~Luo, J.~Li, Y.~Ding, J.~Xie, and J.~Yang, ``Diff-reg: diffusion model in doubly stochastic matrix space for registration problem.''\hskip 1em plus 0.5em minus 0.4em\relax Springer, 2025, pp. 160--178.

\bibitem{zhang2022pcr}
Y.~Zhang, J.~Yu, X.~Huang, W.~Zhou, and J.~Hou, ``Pcr-cg: Point cloud registration via deep explicit color and geometry,'' in \emph{Proc. Eur. Conf. Comput. Vis.}\hskip 1em plus 0.5em minus 0.4em\relax Springer, 2022, pp. 443--459.

\bibitem{huang2022imfnet}
X.~Huang, W.~Qu, Y.~Zuo, Y.~Fang, and X.~Zhao, ``Imfnet: Interpretable multimodal fusion for point cloud registration,'' \emph{IEEE Robotics Autom. Lett.}, vol.~7, no.~4, pp. 12\,323--12\,330, 2022.

\bibitem{chen2022imlovenet}
H.~Chen, Z.~Wei, Y.~Xu, M.~Wei, and J.~Wang, ``Imlovenet: Misaligned image-supported registration network for low-overlap point cloud pairs,'' in \emph{Proc. ACM SIGGRAPH}, 2022, pp. 1--9.

\bibitem{huang2022gmf}
X.~Huang, W.~Qu, Y.~Zuo, Y.~Fang, and X.~Zhao, ``Gmf: General multimodal fusion framework for correspondence outlier rejection,'' \emph{IEEE Robotics Autom. Lett.}, vol.~7, no.~4, pp. 12\,585--12\,592, 2022.

\bibitem{xu2024igreg}
Z.~Xu, X.~Jiang, X.~Gao, R.~Gao, C.~Gu, Q.~Zhang, W.~Li, and X.~Gao, ``Igreg: Image-geometry-assisted point cloud registration via selective correlation fusion,'' \emph{IEEE Trans. Multimed.}, vol.~26, pp. 7475--7489, 2024.

\bibitem{fung2025semreg}
S.~Fung, X.~Lu, D.~d.~S. Edirimuni, W.~Pan, X.~Liu, and H.~Li, ``Semreg: Semantics constrained point cloud registration,'' in \emph{Proc. Eur. Conf. Comput. Vis.}\hskip 1em plus 0.5em minus 0.4em\relax Springer, 2025, pp. 293--310.

\bibitem{yuan2023boosting}
M.~Yuan, X.~Huang, K.~Fu, Z.~Li, and M.~Wang, ``Boosting 3d point cloud registration by transferring multi-modality knowledge,'' in \emph{Proc. IEEE Int. Conf. Robot. Autom.}\hskip 1em plus 0.5em minus 0.4em\relax IEEE, 2023, pp. 11\,734--11\,741.

\bibitem{wang2023zero}
W.~Wang, G.~Mei, B.~Ren, X.~Huang, F.~Poiesi, L.~Van~Gool, N.~Sebe, and B.~Lepri, ``Zero-shot point cloud registration,'' \emph{arXiv preprint arXiv:2312.03032}, 2023.

\bibitem{chen2024pointreggpt}
S.~Chen, H.~Xu, H.~Li, K.~Luo, G.~Liu, C.-W. Fu, P.~Tan, and S.~Liu, ``Pointreggpt: Boosting 3d point cloud registration using generative point-cloud pairs for training,'' \emph{Proc. Eur. Conf. Comput. Vis.}, 2024.

\bibitem{yuan2022pointclm}
M.~Yuan, Z.~Li, Q.~Jin, X.~Chen, and M.~Wang, ``Pointclm: A contrastive learning-based framework for multi-instance point cloud registration,'' in \emph{Proc. Eur. Conf. Comput. Vis.}\hskip 1em plus 0.5em minus 0.4em\relax Springer, 2022, pp. 595--611.

\bibitem{shao2022scrnet}
H.~Shao, Z.~Zhang, X.~Feng, and D.~Zeng, ``Scrnet: A spatial consistency guided network using contrastive learning for point cloud registration,'' \emph{Symmetry}, vol.~14, no.~1, p. 140, 2022.

\bibitem{liu2023density}
Q.~Liu, H.~Zhu, Y.~Zhou, H.~Li, S.~Chang, and M.~Guo, ``Density-invariant features for distant point cloud registration,'' in \emph{Proc. IEEE Int. Conf. Comput. Vis.}, 2023, pp. 18\,215--18\,225.

\bibitem{haitman2025umeregrobust}
Y.~Haitman, A.~Efraim, and J.~M. Francos, ``Uneregrobust-universal manifold embedding compatible features for robust point cloud registration,'' in \emph{Proc. Eur. Conf. Comput. Vis.}\hskip 1em plus 0.5em minus 0.4em\relax Springer, 2025, pp. 358--374.

\bibitem{hatem2023point}
A.~Hatem, Y.~Qian, and Y.~Wang, ``Point-tta: Test-time adaptation for point cloud registration using multitask meta-auxiliary learning,'' in \emph{Proc. Int. Conf. Comput. Vis.}, 2023, pp. 16\,494--16\,504.

\bibitem{wang20203d}
L.~Wang, Y.~Hao, X.~Li, and Y.~Fang, ``3d meta-registration: Learning to learn registration of 3d point clouds,'' \emph{arXiv preprint arXiv:2010.11504}, 2020.

\bibitem{bauer2021reagent}
D.~Bauer, T.~Patten, and M.~Vincze, ``Reagent: Point cloud registration using imitation and reinforcement learning,'' in \emph{Proc. IEEE Conf. Comput. Vis. Pattern Recognit.}, 2021, pp. 14\,586--14\,594.

\bibitem{chen2023point2}
B.~Chen, ``Point cloud registration via heuristic reward reinforcement learning,'' \emph{Stats}, vol.~6, no.~1, pp. 268--278, 2023.

\bibitem{choy2019fully}
C.~Choy, J.~Park, and V.~Koltun, ``Fully convolutional geometric features,'' in \emph{Proc. IEEE Int. Conf. Comput. Vis.}, 2019, pp. 8958--8966.

\bibitem{choy2020deep}
C.~Choy, W.~Dong, and V.~Koltun, ``Deep global registration,'' in \emph{Proc. IEEE Conf. Comput. Vis. Pattern Recognit.}, 2020, pp. 2514--2523.

\bibitem{pais20203dregnet}
G.~D. Pais, S.~Ramalingam, V.~M. Govindu, J.~C. Nascimento, R.~Chellappa, and P.~Miraldo, ``3dregnet: A deep neural network for 3d point registration,'' in \emph{Proc. IEEE Conf. Comput. Vis. Pattern Recognit.}, 2020, pp. 7193--7203.

\bibitem{cao2021pcam}
A.-Q. Cao, G.~Puy, A.~Boulch, and R.~Marlet, ``Pcam: Product of cross-attention matrices for rigid registration of point clouds,'' in \emph{Proc. IEEE Int. Conf. Comput. Vis.}, 2021, pp. 13\,229--13\,238.

\bibitem{wu2023sacf}
Y.~Wu, X.~Hu, Y.~Zhang, M.~Gong, W.~Ma, and Q.~Miao, ``Sacf-net: Skip-attention based correspondence filtering network for point cloud registration,'' \emph{IEEE Trans. Circuits Syst. Video Technol.}, 2023.

\bibitem{wu2023accelerating}
Q.~Wu, J.~Wang, Y.~Zhang, H.~Dong, and C.~Yi, ``Accelerating point cloud registration with low overlap using graphs and sparse convolutions,'' \emph{IEEE Trans. Multimed.}, pp. 1--10, 2023.

\bibitem{yew2022regtr}
Z.~Yew and G.~Lee, ``Regtr: End-to-end point cloud correspondences with transformers,'' in \emph{Proc. IEEE Conf. Comput. Vis. Pattern Recognit.}, 2022, pp. 6677--6686.

\bibitem{yu2023peal}
J.~Yu, L.~Ren, Y.~Zhang, W.~Zhou, L.~Lin, and G.~Dai, ``Peal: Prior-embedded explicit attention learning for low-overlap point cloud registration,'' in \emph{Proc. IEEE Conf. Comput. Vis. Pattern Recognit.}, 2023, pp. 17\,702--17\,711.

\bibitem{liu2023regformer}
J.~Liu, G.~Wang, Z.~Liu, C.~Jiang, M.~Pollefeys, and H.~Wang, ``Regformer: an efficient projection-aware transformer network for large-scale point cloud registration,'' in \emph{Proc. IEEE Int. Conf. Comput. Vis.}, 2023, pp. 8451--8460.

\bibitem{yuan2023egst}
Y.~Yuan, Y.~Wu, X.~Fan, M.~Gong, W.~Ma, and Q.~Miao, ``Egst: Enhanced geometric structure transformer for point cloud registration,'' \emph{IEEE Trans. Vis. Comput. Graph.}, vol.~30, no.~9, pp. 6222--6234, 2023.

\bibitem{xiong2024speal}
K.~Xiong, M.~Zheng, Q.~Xu, C.~Wen, S.~Shen, and C.~Wang, ``Speal: Skeletal prior embedded attention learning for cross-source point cloud registration,'' in \emph{Proc. AAAI Conf. Artif. Intell.}, vol.~38, no.~6, 2024, pp. 6279--6287.

\bibitem{wang2024neighborhood}
Y.~Wang, P.~Zhou, G.~Geng, L.~An, K.~Li, and R.~Li, ``Neighborhood multi-compound transformer for point cloud registration,'' \emph{IEEE Trans. Circuits Syst. Video Technol.}, vol.~34, no.~9, pp. 8469--8480, 2024.

\bibitem{chen2024dynamic}
H.~Chen, P.~Yan, S.~Xiang, and Y.~Tan, ``Dynamic cues-assisted transformer for robust point cloud registration,'' in \emph{Proc. IEEE Conf. Comput. Vis. Pattern Recognit.}, 2024, pp. 21\,698--21\,707.

\bibitem{yu2024learning}
Z.~Yu, Z.~Qin, L.~Zheng, and K.~Xu, ``Learning instance-aware correspondences for robust multi-instance point cloud registration in cluttered scenes,'' in \emph{Proc. IEEE Conf. Comput. Vis. Pattern Recognit.}, 2024, pp. 19\,605--19\,614.

\bibitem{wang2019deep}
Y.~Wang and J.~M. Solomon, ``Deep closest point: Learning representations for point cloud registration,'' in \emph{Proc. IEEE Int. Conf. Comput. Vis.}, 2019, pp. 3523--3532.

\bibitem{li2020iterative}
J.~Li, C.~Zhang, Z.~Xu, H.~Zhou, and C.~Zhang, ``Iterative distance-aware similarity matrix convolution with mutual-supervised point elimination for efficient point cloud registration,'' in \emph{Proc. Eur. Conf. Comput. Vis.}, 2020, pp. 378--394.

\bibitem{wiesmann2022dcpcr}
L.~Wiesmann, T.~Guadagnino, I.~Vizzo, G.~Grisetti, J.~Behley, and C.~Stachniss, ``Dcpcr: Deep compressed point cloud registration in large-scale outdoor environments,'' \emph{IEEE Robot. Autom. Lett.}, vol.~7, no.~3, pp. 6327--6334, 2022.

\bibitem{zheng2022global}
Y.~Zheng, Y.~Li, S.~Yang, and H.~Lu, ``Global-pbnet: A novel point cloud registration for autonomous driving,'' \emph{IEEE Trans. Intell. Transp. Syst.}, vol.~23, no.~11, pp. 22\,312--22\,319, 2022.

\bibitem{yew2020rpm}
Z.~J. Yew and G.~H. Lee, ``Rpm-net: Robust point matching using learned features,'' in \emph{Proc. IEEE Conf. Comput. Vis. Pattern Recognit.}, 2020, pp. 11\,824--11\,833.

\bibitem{aoki2019pointnetlk}
Y.~Aoki, H.~Goforth, R.~Srivatsan, and L., ``Pointnetlk: Robust \& efficient point cloud registration using pointnet,'' in \emph{Proc. IEEE Conf. Comput. Vis. Pattern Recognit.}, 2019, pp. 7163--7172.

\bibitem{li2021pointnetlk}
X.~Li, J.~K. Pontes, and S.~Lucey, ``Pointnetlk revisited,'' in \emph{Proc. IEEE Conf. Comput. Vis. Pattern Recognit.}, 2021, pp. 12\,763--12\,772.

\bibitem{zhang2022vrnet}
Z.~Zhang, J.~Sun, Y.~Dai, B.~Fan, and M.~He, ``Vrnet: Learning the rectified virtual corresponding points for 3d point cloud registration,'' \emph{IEEE Trans. Circuits Syst. Video Technol.}, vol.~32, no.~8, pp. 4997--5010, 2022.

\bibitem{qi2017pointnet}
C.~R. Qi, H.~Su, K.~Mo, and L.~J. Guibas, ``Pointnet: Deep learning on point sets for 3d classification and segmentation,'' in \emph{Proc. IEEE Conf. Comput. Vis. Pattern Recognit.}, 2017, pp. 652--660.

\bibitem{qi2017pointnet++}
C.~R. Qi, L.~Yi, H.~Su, and L.~J. Guibas, ``Pointnet++: Deep hierarchical feature learning on point sets in a metric space,'' \emph{Proc. Int. Conf. Neural Inf. Process. Syst.}, vol.~30, 2017.

\bibitem{he2016deep}
K.~He, X.~Zhang, S.~Ren, and J.~Sun, ``Deep residual learning for image recognition,'' in \emph{Proc. IEEE Conf. Comput. Vis. Pattern Recognit.}, 2016, pp. 770--778.

\bibitem{wu2022inenet}
Y.~Wu, Y.~Zhang, X.~Fan, M.~Gong, Q.~Miao, and W.~Ma, ``Inenet: Inliers estimation network with similarity learning for partial overlapping registration,'' \emph{IEEE Trans. Circuits Syst. Video Technol.}, vol.~33, no.~3, pp. 1413--1426, 2022.

\bibitem{vaswani2017attention}
A.~Vaswani, N.~Shazeer, N.~Parmar, J.~Uszkoreit, L.~Jones, A.~Gomez, {\L}.~Kaiser, and I.~Polosukhin, ``Attention is all you need,'' in \emph{Proc. Int. Conf. Neural Inf. Process. Syst.}, 2017.

\bibitem{jang2016categorical}
E.~Jang, S.~Gu, and B.~Poole, ``Categorical reparameterization with gumbel-softmax,'' \emph{arXiv preprint arXiv:1611.01144}, 2016.

\bibitem{wang2019dynamic}
Y.~Wang, Y.~Sun, Z.~Liu, S.~E. Sarma, M.~M. Bronstein, and J.~M. Solomon, ``Dynamic graph cnn for learning on point clouds,'' \emph{ACM Trans. Graph.}, vol.~38, no.~5, pp. 1--12, 2019.

\bibitem{sinkhorn1964relationship}
R.~Sinkhorn, ``A relationship between arbitrary positive matrices and doubly stochastic matrices,'' \emph{Ann. Math. Stat.}, vol.~35, no.~2, pp. 876--879, 1964.

\bibitem{zhao2024sgor}
G.~Zhao, Z.~Guo, and H.~Ma, ``Sgor: Outlier removal by leveraging semantic and geometric information for robust point cloud registration,'' \emph{arXiv preprint arXiv:2407.06297}, 2024.

\bibitem{li2024effective}
R.~Li, X.~Yuan, S.~Gan, R.~Bi, S.~Gao, W.~Luo, and C.~Chen, ``An effective point cloud registration method based on robust removal of outliers,'' \emph{IEEE Trans. Geosci. Remote Sens.}, vol.~62, 2024.

\bibitem{han2024robust}
J.~Han, M.~Shin, and J.~Paik, ``Robust point cloud registration using hough voting-based correspondence outlier rejection,'' \emph{Eng. Appl. Artif. Intell.}, vol. 133, p. 107985, 2024.

\bibitem{ma2024pcgor}
G.~Ma, H.~Wei, R.~Lin, and J.~Wu, ``Pcgor: A novel plane constraints-based guaranteed outlier removal method for large-scale lidar point cloud registration,'' \emph{IEEE Trans. Geosci. Remote Sens.}, 2024.

\bibitem{min2021convolutional}
J.~Min and M.~Cho, ``Convolutional hough matching networks,'' in \emph{Proc. IEEE Conf. Comput. Vis. Pattern Recognit.}, 2021, pp. 2940--2950.

\bibitem{zhang20233d}
X.~Zhang, J.~Yang, S.~Zhang, and Y.~Zhang, ``3d registration with maximal cliques,'' in \emph{Proc. IEEE Conf. Comput. Vis. Pattern Recognit.}, 2023, pp. 17\,745--17\,754.

\bibitem{fischler1981random}
M.~Fischler and R.~Bolles, ``Random sample consensus: a paradigm for model fitting with applications to image analysis and automated cartography,'' \emph{Communications of the ACM}, vol.~24, no.~6, pp. 381--395, 1981.

\bibitem{zhang2022point}
Y.~Zhang, Z.~Sun, Z.~Zeng, and K.~Lam, ``Point cloud registration using multiattention mechanism and deep hybrid features,'' \emph{IEEE Intell. Syst.}, vol.~38, no.~1, pp. 58--68, 2022.

\bibitem{arun1987least}
K.~Arun, T.~Huang, and S.~Blostein, ``Least-squares fitting of two 3-d point sets,'' \emph{IEEE Trans. Pattern Anal. Mach. Intell.}, no.~5, pp. 698--700, 1987.

\bibitem{deng2018ppf}
H.~Deng, T.~Birdal, and S.~Ilic, ``Ppf-foldnet: Unsupervised learning of rotation invariant 3d local descriptors,'' in \emph{Proc. Eur. Conf. Comput. Vis.}, 2018, pp. 602--618.

\bibitem{li2024dbdnet}
S.~Li, J.~Zhu, and Y.~Xie, ``Dbdnet: Partial-to-partial point cloud registration with dual branches decoupling,'' \emph{Knowl.-Based Syst.}, vol. 296, p. 111864, 2024.

\bibitem{newman1993model}
T.~S. Newman, P.~J. Flynn, and A.~K. Jain, ``Model-based classification of quadric surfaces,'' \emph{CVGIP: Image Understand.}, vol.~58, no.~2, pp. 235--249, 1993.

\bibitem{zhang2023pyrf}
J.~Zhang, S.~Huang, J.~Liu, X.~Zhu, and F.~Xu, ``Pyrf-pcr: A robust three-stage 3d point cloud registration for outdoor scene,'' \emph{IEEE Trans. Intell. Vehicles}, vol.~9, pp. 1270--1281, 2023.

\bibitem{eckart2018hgmr}
B.~Eckart, K.~Kim, and J.~Kautz, ``Hgmr: Hierarchical gaussian mixtures for adaptive 3d registration,'' in \emph{Proc. Eur. Conf. Comput. Vis.}, 2018, pp. 705--721.

\bibitem{hertz2020pointgmm}
A.~Hertz, R.~Hanocka, R.~Giryes, and D.~Cohen-Or, ``Pointgmm: A neural gmm network for point clouds,'' in \emph{Proc. IEEE Conf. Comput. Vis. Pattern Recognit.}, 2020, pp. 12\,054--12\,063.

\bibitem{chen2019clusternet}
C.~Chen, G.~Li, R.~Xu, T.~Chen, M.~Wang, and L.~Lin, ``Clusternet: Deep hierarchical cluster network with rigorously rotation-invariant representation for point cloud analysis,'' in \emph{Proc. IEEE Conf. Comput. Vis. Pattern Recognit.}, 2019, pp. 4994--5002.

\bibitem{sohl2015deep}
J.~Sohl-Dickstein, E.~Weiss, N.~Maheswaranathan, and S.~Ganguli, ``Deep unsupervised learning using nonequilibrium thermodynamics,'' in \emph{Proc. Int. Conf. Mach. Learn.}\hskip 1em plus 0.5em minus 0.4em\relax PMLR, 2015, pp. 2256--2265.

\bibitem{ho2020denoising}
J.~Ho, A.~Jain, and P.~Abbeel, ``Denoising diffusion probabilistic models,'' \emph{Proc. Int. Conf. Neural Inf. Process. Syst.}, vol.~33, pp. 6840--6851, 2020.

\bibitem{luo2021diffusion}
S.~Luo and W.~Hu, ``Diffusion probabilistic models for 3d point cloud generation,'' in \emph{Proc. IEEE Conf. Comput. Vis. Pattern Recognit.}, 2021, pp. 2837--2845.

\bibitem{gong2024eadreg}
L.~Gong, J.~Liu, J.~Ma, L.~Liu, Y.~Wang, and H.~Wang, ``Eadreg: Probabilistic correspondence generation with efficient autoregressive diffusion model for outdoor point cloud registration,'' \emph{arXiv preprint arXiv:2411.15271}, 2024.

\bibitem{chamberlain2021grand}
B.~Chamberlain, J.~Rowbottom, M.~I. Gorinova, M.~Bronstein, S.~Webb, and E.~Rossi, ``Grand: Graph neural diffusion,'' in \emph{Int. Conf. Mach. Learn.}\hskip 1em plus 0.5em minus 0.4em\relax PMLR, 2021, pp. 1407--1418.

\bibitem{sun2009concise}
J.~Sun, M.~Ovsjanikov, and L.~Guibas, ``A concise and provably informative multi-scale signature based on heat diffusion,'' in \emph{Comput. Graph. Forum}, vol.~28, no.~5.\hskip 1em plus 0.5em minus 0.4em\relax Wiley Online Library, 2009, pp. 1383--1392.

\bibitem{shoemake1985animating}
K.~Shoemake, ``Animating rotation with quaternion curves,'' in \emph{Proc. Conf. Comput. Graph. Interact. Techn.}, 1985, pp. 245--254.

\bibitem{xie2023cross}
Y.~Xie, J.~Zhu, S.~Li, and P.~Shi, ``Cross-modal information-guided network using contrastive learning for point cloud registration,'' \emph{IEEE Robotics and Automation Letters}, vol.~9, no.~1, pp. 103--110, 2023.

\bibitem{zheng2024point}
X.~Zheng, X.~Huang, G.~Mei, Y.~Hou, Z.~Lyu, B.~Dai, W.~Ouyang, and Y.~Gong, ``Point cloud pre-training with diffusion models,'' in \emph{Proc. IEEE Conf. Comput. Vis. Pattern Recognit.}, 2024, pp. 22\,935--22\,945.

\bibitem{yamada2022point}
R.~Yamada, H.~Kataoka, N.~Chiba, Y.~Domae, and T.~Ogata, ``Point cloud pre-training with natural 3d structures,'' in \emph{Proc. IEEE Conf. Comput. Vis. Pattern Recognit.}, 2022, pp. 21\,283--21\,293.

\bibitem{efraim2022estimating}
A.~Efraim and J.~M. Francos, ``Estimating rigid transformations of noisy point clouds using the universal manifold embedding,'' \emph{J. Math. Imaging Vis.}, vol.~64, no.~4, pp. 343--363, 2022.

\bibitem{hagege2016universal}
R.~R. Hagege and J.~M. Francos, ``Universal manifold embedding for geometrically deformed functions,'' \emph{IEEE Trans. Inf. Theory}, vol.~62, no.~6, pp. 3676--3684, 2016.

\bibitem{choy20194d}
C.~Choy, J.~Gwak, and S.~Savarese, ``4d spatio-temporal convnets: Minkowski convolutional neural networks,'' in \emph{Proc. IEEE Conf. Comput. Vis. Pattern Recognit.}, 2019, pp. 3075--3084.

\bibitem{niu2021review}
Z.~Niu, G.~Zhong, and H.~Yu, ``A review on the attention mechanism of deep learning,'' \emph{Neurocomputing}, vol. 452, pp. 48--62, 2021.

\bibitem{zhao2023patch}
T.~Zhao, L.~Li, T.~Tian, J.~Ma, and J.~Tian, ``Patch-guided point matching for point cloud registration with low overlap,'' \emph{Pattern Recognit.}, vol. 144, p. 109876, 2023.

\bibitem{wang2023ccag}
Y.~Wang, P.~Zhou, G.~Geng, L.~An, and Y.~Liu, ``Ccag: End-to-end point cloud registration,'' \emph{IEEE Robot. Autom. Lett.}, vol.~9, no.~1, pp. 435--442, 2023.

\bibitem{yu2021cofinet}
H.~Yu, F.~Li, M.~Saleh, B.~Busam, and S.~Ilic, ``Cofinet: Reliable coarse-to-fine correspondences for robust pointcloud registration,'' \emph{Proc. Int. Conf. Neural Inf. Process. Syst.}, vol.~34, pp. 23\,872--23\,884, 2021.

\bibitem{yu2024riga}
H.~Yu, J.~Hou, Z.~Qin, M.~Saleh, I.~Shugurov, K.~Wang, B.~Busam, and S.~Ilic, ``Riga: Rotation-invariant and globally-aware descriptors for point cloud registration,'' \emph{IEEE Trans. Pattern Anal. Mach. Intell.}, vol.~46, no.~5, pp. 3796--3812, 2024.

\bibitem{li2022lepard}
Y.~Li and T.~Harada, ``Lepard: Learning partial point cloud matching in rigid and deformable scenes,'' in \emph{IEEE Conf. Comput. Vis. Pattern Recognit.}, 2022, pp. 5554--5564.

\bibitem{wang2024point}
M.~Wang, G.~Chen, Y.~Yang, L.~Yuan, and Y.~Yue, ``Point tree transformer for point cloud registration,'' \emph{arXiv preprint arXiv:2406.17530}, 2024.

\bibitem{zhao2024lfa}
Z.~Zhao, J.~Kang, L.~Feng, J.~Liang, Y.~Ren, and B.~Wu, ``Lfa-net: Enhanced pointnet and assignable weights transformer network for partial-to-partial point cloud registration,'' \emph{IEEE Trans. Circuits Syst. Video Technol.}, 2024.

\bibitem{horache20213d}
S.~Horache, J.-E. Deschaud, and F.~Goulette, ``3d point cloud registration with multi-scale architecture and unsupervised transfer learning,'' in \emph{Proc. Int. Conf. 3D Vis.}\hskip 1em plus 0.5em minus 0.4em\relax IEEE, 2021, pp. 1351--1361.

\bibitem{zhang2021representation}
Z.~Zhang, J.~Sun, Y.~Dai, D.~Zhou, X.~Song, and M.~He, ``A representation separation perspective to correspondence-free unsupervised 3-d point cloud registration,'' \emph{IEEE Geosci. Remote Sens. Lett.}, vol.~19, pp. 1--5, 2021.

\bibitem{wang2020unsupervised}
L.~Wang, X.~Li, and Y.~Fang, ``Unsupervised learning of 3d point set registration,'' \emph{arXiv preprint arXiv:2006.06200}, 2020.

\bibitem{sun2023research}
X.~J. Sun, W.~H. Li, J.~W. Huang, D.~Y. Chen, and T.~Jia, ``Research and application on cross-source point cloud registration method based on unsupervised learning,'' in \emph{Proc. Int. Conf. CYBER Technol. Autom., Control, Intell. Syst.}\hskip 1em plus 0.5em minus 0.4em\relax IEEE, 2023, pp. 1327--1332.

\bibitem{yuan2024MMIPCR}
Y.~Yuan, Y.~Wu, M.~Yue, M.~Gong, X.~Fan, W.~Ma, and Q.~Miao, ``Learning discriminative features via multi-hierarchical mutual information for unsupervised point cloud registration,'' \emph{IEEE Trans. Circuits Syst. Video Technol.}, 2024.

\bibitem{wang2022improving}
Z.~Wang, X.~Huo, Z.~Chen, J.~Zhang, L.~Sheng, and D.~Xu, ``Improving rgb-d point cloud registration by learning multi-scale local linear transformation,'' in \emph{Proc. Eur. Conf. Comput. Vis.}\hskip 1em plus 0.5em minus 0.4em\relax Springer, 2022, pp. 175--191.

\bibitem{yuan2023pointmbf}
M.~Yuan, K.~Fu, Z.~Li, Y.~Meng, and M.~Wang, ``Pointmbf: A multi-scale bidirectional fusion network for unsupervised rgb-d point cloud registration,'' in \emph{Proc. IEEE Int. Conf. Comput. Vis.}, 2023, pp. 17\,694--17\,705.

\bibitem{yan2024discriminative}
C.~Yan, M.~Feng, Z.~Wu, Y.~Guo, W.~Dong, Y.~Wang, and A.~Mian, ``Discriminative correspondence estimation for unsupervised rgb-d point cloud registration,'' \emph{IEEE Trans. Circuits Syst. Video Technol.}, 2024.

\bibitem{yu2024nerf}
Z.~Yu, Z.~Qin, Y.~Tang, Y.~Wang, R.~Yi, C.~Zhu, and K.~Xu, ``Nerf-guided unsupervised learning of rgb-d registration,'' \emph{arXiv preprint arXiv:2405.00507}, 2024.

\bibitem{jiang2021planning}
H.~Jiang, J.~Xie, J.~Qian, and J.~Yang, ``Planning with learned dynamic model for unsupervised point cloud registration,'' \emph{Proc. Int. Joint Conf. Artif. Intell.}, pp. 772--778, 2021.

\bibitem{jiang2021sampling}
H.~Jiang, Y.~Shen, J.~Xie, J.~Li, J.~Qian, and J.~Yang, ``Sampling network guided cross-entropy method for unsupervised point cloud registration,'' in \emph{Proc. IEEE Int. Conf. Comput. Vis.}, 2021, pp. 6128--6137.

\bibitem{shi2023overlap}
P.~Shi, J.~Zhang, H.~Cheng, J.~Wang, Y.~Zhou, C.~Zhao, and J.~Zhu, ``Overlap bias matching is necessary for point cloud registration,'' \emph{arXiv preprint arXiv:2308.09364}, 2023.

\bibitem{lang2021deepume}
N.~Lang and J.~M. Francos, ``Deepume: Learning the universal manifold embedding for robust point cloud registration,'' \emph{arXiv preprint arXiv:2112.09938}, 2021.

\bibitem{zeng2021corrnet3d}
Y.~Zeng, Y.~Qian, Z.~Zhu, J.~Hou, H.~Yuan, and Y.~He, ``Corrnet3d: Unsupervised end-to-end learning of dense correspondence for 3d point clouds,'' in \emph{Proc. IEEE Conf. Comput. Vis. Pattern Recognit.}, 2021, pp. 6052--6061.

\bibitem{kadam2022r}
P.~Kadam, M.~Zhang, S.~Liu, and C.-C.~J. Kuo, ``R-pointhop: A green, accurate, and unsupervised point cloud registration method,'' \emph{IEEE Trans. Image Process.}, vol.~31, pp. 2710--2725, 2022.

\bibitem{jiang2024gtinet}
Y.~Jiang, B.~Zhou, X.~Liu, Q.~Li, and C.~Cheng, ``Gtinet: Global topology-aware interactions for unsupervised point cloud registration,'' \emph{IEEE Trans. Circuits Syst. Video Technol.}, 2024.

\bibitem{shen2022reliable}
Y.~Shen, L.~Hui, H.~Jiang, J.~Xie, and J.~Yang, ``Reliable inlier evaluation for unsupervised point cloud registration,'' in \emph{Proc. AAAI Conf. Artif. Intell.}, vol.~36, no.~2, 2022, pp. 2198--2206.

\bibitem{yuan2024inlier}
Y.~Yuan, Y.~Wu, X.~Fan, M.~Gong, Q.~Miao, and W.~Ma, ``Inlier confidence calibration for point cloud registration,'' in \emph{Proc. IEEE Conf. Comput. Vis. Pattern Recognit.}, 2024, pp. 5312--5321.

\bibitem{zheng2024regiformer}
C.~Zheng, M.~Ma, Z.~Chen, H.~Chen, W.~Wang, and M.~Wei, ``Regiformer: Unsupervised point cloud registration via geometric local-to-global transformer and self augmentation,'' \emph{IEEE Trans. Geosci. Remote Sens.}, vol.~62, 2024.

\bibitem{liu2024extend}
Q.~Liu, H.~Zhu, Z.~Wang, Y.~Zhou, S.~Chang, and M.~Guo, ``Extend your own correspondences: Unsupervised distant point cloud registration by progressive distance extension,'' in \emph{Proc. IEEE Conf. Comput. Vis. Pattern Recognit.}, 2024, pp. 20\,816--20\,826.

\bibitem{xiong2024mining}
K.~Xiong, H.~Xiang, Q.~Xu, C.~Wen, S.~Shen, J.~Li, and C.~Wang, ``Mining and transferring feature-geometry coherence for unsupervised point cloud registration,'' \emph{Proc. Int. Conf. Neural Inf. Process. Syst.}, 2024.

\bibitem{segal2009generalized}
A.~Segal, D.~Haehnel, and S.~Thrun, ``Generalized-icp.'' in \emph{Robotics: Sci. Syst.}, vol.~2, no.~4.\hskip 1em plus 0.5em minus 0.4em\relax Seattle, WA, 2009, p. 435.

\bibitem{narendra1977branch}
Narendra and Fukunaga, ``A branch and bound algorithm for feature subset selection,'' \emph{IEEE Trans. Comput.}, vol. 100, no.~9, pp. 917--922, 1977.

\bibitem{gold1998new}
S.~Gold, A.~Rangarajan, C.-P. Lu, S.~Pappu, and E.~Mjolsness, ``New algorithms for 2d and 3d point matching: Pose estimation and correspondence,'' \emph{Pattern Recognit.}, vol.~31, no.~8, pp. 1019--1031, 1998.

\bibitem{lucas1981iterative}
B.~D. Lucas and T.~Kanade, ``An iterative image registration technique with an application to stereo vision,'' in \emph{Proc. Int. Joint Conf. Artif. Intell.}, vol.~2, 1981, pp. 674--679.

\bibitem{chen2024rgbd}
C.~Chen, X.~Jia, Y.~Zheng, and Y.~Qu, ``Rgbd-glue: General feature combination for robust rgb-d point cloud registration,'' \emph{arXiv preprint arXiv:2405.07594}, 2024.

\bibitem{mildenhall2021nerf}
B.~Mildenhall, P.~P. Srinivasan, M.~Tancik, J.~T. Barron, R.~Ramamoorthi, and R.~Ng, ``Nerf: Representing scenes as neural radiance fields for view synthesis,'' \emph{Commun. ACM}, vol.~65, no.~1, pp. 99--106, 2021.

\bibitem{efraim2019universal}
A.~Efraim and J.~M. Francos, ``The universal manifold embedding for estimating rigid transformations of point clouds,'' in \emph{Proc. IEEE Int. Conf. Acoust. Speech Signal Process.}\hskip 1em plus 0.5em minus 0.4em\relax IEEE, 2019, pp. 5157--5161.

\bibitem{radford2021learning}
A.~Radford, J.~W. Kim, C.~Hallacy, A.~Ramesh, G.~Goh, S.~Agarwal, G.~Sastry, A.~Askell, P.~Mishkin, J.~Clark \emph{et~al.}, ``Learning transferable visual models from natural language supervision,'' in \emph{Proc. Int. Conf. Mach. Learn}.\hskip 1em plus 0.5em minus 0.4em\relax PMLR, 2021, pp. 8748--8763.

\bibitem{zhang2022pointclip}
R.~Zhang, Z.~Guo, W.~Zhang, K.~Li, X.~Miao, B.~Cui, Y.~Qiao, P.~Gao, and H.~Li, ``Pointclip: Point cloud understanding by clip,'' in \emph{Proc. IEEE Conf. Comput. Vis. Pattern Recognit.}, 2022, pp. 8552--8562.

\bibitem{huang2023clip2point}
T.~Huang, B.~Dong, Y.~Yang, X.~Huang, R.~W. Lau, W.~Ouyang, and W.~Zuo, ``Clip2point: Transfer clip to point cloud classification with image-depth pre-training,'' in \emph{Proc. IEEE Int. Conf. Comput. Vis.}, 2023, pp. 22\,157--22\,167.

\bibitem{zeng2023clip2}
Y.~Zeng, C.~Jiang, J.~Mao, J.~Han, C.~Ye, Q.~Huang, D.-Y. Yeung, Z.~Yang, X.~Liang, and H.~Xu, ``Clip2: Contrastive language-image-point pretraining from real-world point cloud data,'' in \emph{Proc. IEEE Conf. Comput. Vis. Pattern Recognit.}, 2023, pp. 15\,244--15\,253.

\end{thebibliography}

\end{document}